\pdfoutput=1

\documentclass[11pt]{article}

\usepackage[]{acl}

\usepackage{times}
\usepackage{latexsym}

\usepackage[T1]{fontenc}

\usepackage[utf8]{inputenc}

\usepackage{microtype}

\usepackage{inconsolata}

\usepackage{booktabs}
\usepackage{array}
\usepackage{xcolor,colortbl}
\definecolor{orange}{rgb}{1,0.647,0}
\definecolor{grey}{rgb}{0.5,0.5,0.5}
\definecolor{red}{rgb}{1,0,0}
\definecolor{blue}{rgb}{0,0,1}

\pdfoutput=1

\usepackage{dialogue}
\usepackage{pifont}
\usepackage{tabularx}
\usepackage{booktabs} 
\usepackage[T1]{fontenc}

\usepackage[utf8]{inputenc}

\usepackage{microtype}

\usepackage{inconsolata}
\usepackage{algorithm}
\usepackage{algorithmic}

\pdfoutput=1
\usepackage{graphicx} 
\usepackage[T1]{fontenc}

\usepackage{times}
\usepackage{latexsym}
\usepackage{caption}
\usepackage{graphicx}
\usepackage{multirow}
\usepackage{times}

\usepackage[utf8]{inputenc}

\usepackage{microtype}

\usepackage{inconsolata}
\usepackage{tabularx}

\usepackage{array}
\usepackage{booktabs}
\usepackage[capitalize,noabbrev]{cleveref}

\newcommand{\mcraft}{MentalAgora}
\definecolor{rowcolor1}{rgb}{0.9, 0.9, 0.9} 
\definecolor{rowcolor2}{rgb}{1, 1, 1} 

\title{\mcraft: A Gateway to Advanced Personalized Care in Mental Health through Multi-Agent Debating and Attribute Control}

\author{Yeonji Lee$^{1}$ \hspace{1cm} Sangjun Park$^{1,2}$ \hspace{1cm} Kyunghyun Cho$^{3^*}$ \hspace{1cm} JinYeong Bak$^{1^*}$\\
  $^1$Sungkyunkwan University, Suwon, South Korea \\
  $^2$Upstage AI, South Korea \\
  $^3$New York University, NY, USA \\
  \texttt{yeonjilee@g.skku.edu}, \texttt{sangjun@upstage.ai}, \\
  \texttt{kc119@nyu.edu}, \texttt{jy.bak@skku.edu} \\
 }

\begin{document}
\maketitle
\begingroup\def\thefootnote{*}\footnotetext{Corresponding authors}\endgroup
\begin{abstract}
As mental health issues globally escalate, there is a tremendous need for advanced digital support systems. We introduce \mcraft, a novel framework employing large language models enhanced by interaction between multiple agents for tailored mental health support. This framework operates through three stages: strategic debating, tailored counselor creation, and response generation, enabling the dynamic customization of responses based on individual user preferences and therapeutic needs.
We conduct experiments utilizing a high-quality evaluation dataset TherapyTalk crafted with mental health professionals, showing that \mcraft{} generates expert-aligned and user preference-enhanced responses.
Our evaluations, including experiments and user studies, demonstrate that \mcraft ~aligns with professional standards and effectively meets user preferences, setting a new benchmark for digital mental health interventions.\footnote{Code and Dataset: \url{https://github.com/jennylee03/MentalAgora}}

\end{abstract}

\section{Introduction}
\label{1-Introduction}
With 1 in every 8 people confirming they have experienced mental disorders~\citep{who2022}, estimates suggest that nearly half of the global population will encounter mental health challenges at some point~\citep{MCGRATH2023668}. 
This trend underscores the need for effective and personalized support systems tailored to meet the needs of individuals~\citep{lu-etal-2023-miracle,chen2023large}. Moreover, the rise of digital platforms has transformed communication, increasing the demand for online support and engagement. This shift highlights the importance of accessible and effective digital interventions.

Due to excellent communication skills, there are ongoing efforts to apply Large Language Models (LLMs) in the fields of sociology, psychology, and psychiatry \citep{sharma-etal-2023-cognitive, chiu2024computational, li-etal-2023-theory}.
Also, to leverage the benefits of collective intelligence, research into imitating human interactions is actively underway~\citep{zhang2024exploring,chan2023chateval} using multi-agent debating and simulation systems.  

In the mental health domain, it is an ongoing topic to provide emotionally supportive responses for people having mental concerns \citep{hua2024large}.
Recent advancements have improved their ability to generate empathetic and clinically relevant responses~\citep{wei2021emotionaware, loh2023harnessing,shen-etal-2020-counseling}.
Despite these improvements, a single LLM still struggles to comprehensively understand users' complex situations, limiting the depth and specificity of responses \citep{guo2024large}. 

Addressing this gap, we propose a new framework, \mcraft{}, adopting a multi-agent method in the mental health domain and supporting response generation. This framework leverages the concept of multi-agent interaction~\cite{chen2023agentverse, park2023generative}, where each agent models different aspects, providing a richer interpretation.
\mcraft{} applies various counseling strategies as attributes of LLM agents, involving them in debating to generate the most suitable persona addressing user concerns. This persona prompt enables LLMs to provide the most appropriate responses for the user.
To verify the effectiveness of \mcraft{}, we create the TherapyTalk dataset. The automatic and human evaluation results show that \mcraft{} outperforms other methods, achieving the highest alignment with expert responses and satisfying user preferences.

Furthermore, we collect actual user concerns to verify whether \mcraft{} can provide assistance to real users, offering responses to their issues. As a result, \mcraft{} demonstrates greater user satisfaction compared to all other baselines, highlighting its potential in digital mental health interventions.

Our contributions in this paper are manifold:
\begin{itemize}
    \item We develop a novel framework, \mcraft, to facilitate dynamic and nuanced interactions via multi-agent debating, delivering mental health support based on user preferences. 
    \item We enhance attribute controllable generation to dynamically customize responses by allowing for a multifaceted counselor persona creation tailored to a wide range of user preferences.
    \item We curate a high-quality dataset, TherapyTalk, in collaboration with mental health professionals to ensure responses are grounded in expert knowledge.
    \item Our work showcases the potential utility of this approach in transforming mental health support, validated through comprehensive experiments and user study.
\end{itemize}

\begin{figure*}[t!]
    \centering
    \includegraphics[width=1\linewidth]{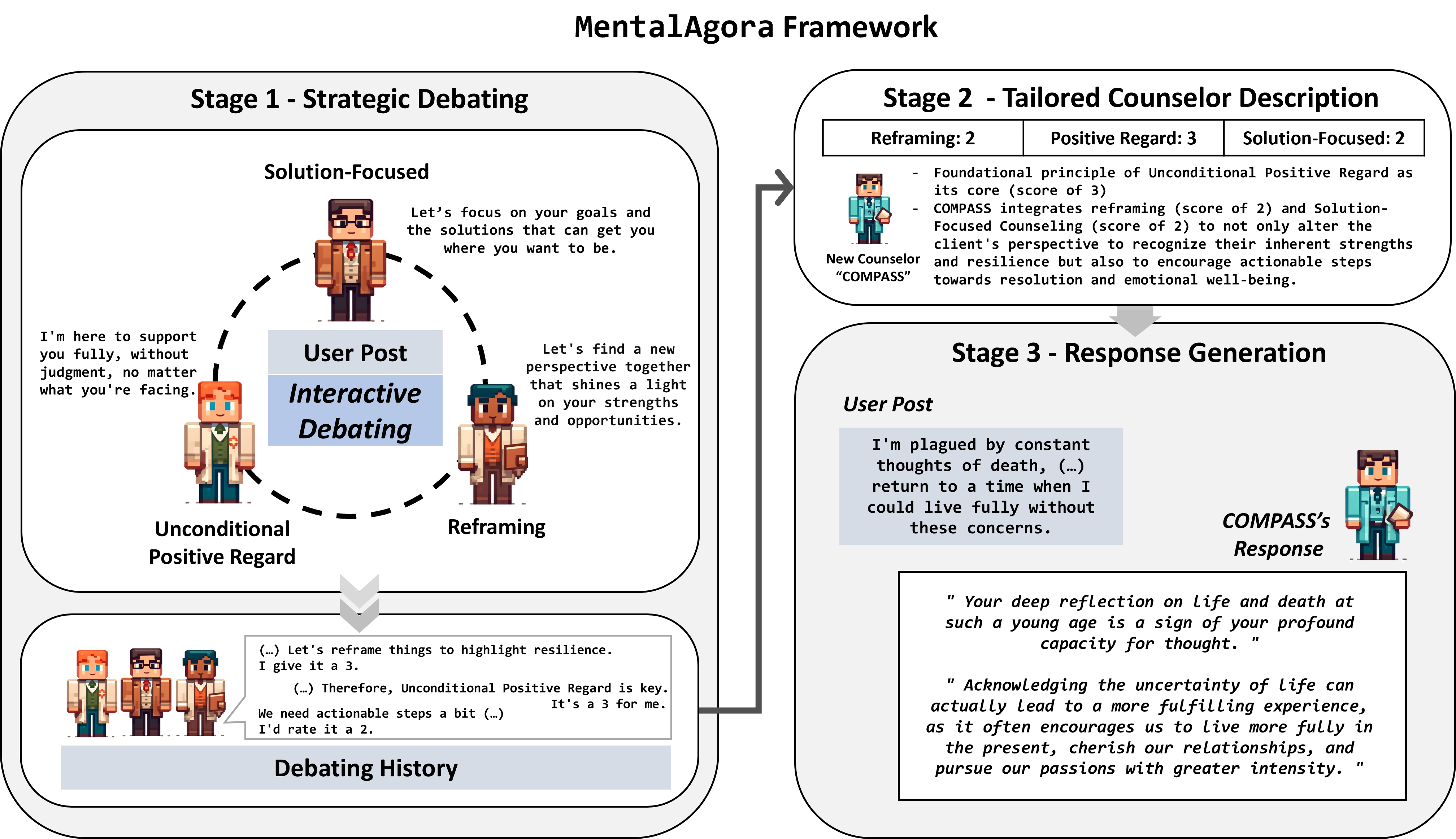}
    \caption{\mcraft{} framework overview. This diagram outlines the \mcraft ~framework, showcasing its three stages: Strategic Debating, Tailored Counselor Creation, and Response Generation, which collectively enhance the creation of personalized therapeutic responses based on user-specific needs.}
    \label{fig:MIND-CRAFT}
\end{figure*}

\section{Related Work}
\label{2-Related-work}
In mental health support response generation, most research has been conducted with single-agent LLMs, which often yield general and static responses.
Multi-agent methods can overcome these limitations, generating more dynamic and nuanced responses.
However, their application in the mental health domain has been limited despite success in evaluation or reasoning.
This is notable, as the intricate nature of mental health issues could benefit from the diverse perspectives and responses offered by multi-agent systems.
                        
\subsection{Mental Health Support Response Generation}
The majority of these studies uses Transformers and LLMs to discern the user's intent and emotions within the conversation. 
Following \citet{rashkin2019empathetic}'s introduction of the empathetic dataset and task, subsequent studies have focused on generating and rewriting empathetic responses~\citep{sharma2021facilitating, mishra-etal-2023-pal, shen-etal-2020-counseling}.
These studies aim to enhance the detection of emotions and sentiments in user utterances~\citep{zhan2022feel} and strive to attain greater empathy.
Building on counselor reflection as a key verbal skill, \citet{min-etal-2022-pair} developed a system for analyzing these reflections, while \citet{wang-etal-2023-cue} leveraged user profiles and comment histories to generate personalized responses, using generative models conditioned on the questions and respondents' data.

\citet{sharma-etal-2023-cognitive} introduced a cognitive therapeutic technique aimed at countering negative thoughts by replacing them with more optimistic reframed thoughts highlighting the capability of LLMs in the psychiatry field. Additionally, \citet{liu2023chatcounselor} discusses the development of chatbots designed to enhance counseling interactions in terms of psychological support. These studies have laid the groundwork for support response generation, setting the stage for our research that incorporates multi-agent systems in this domain.

\subsection{Controlled Text Generation}
In the domain of controlled text generation, \citet{Zou_2021} proposed ``inverse prompting'' to enhance control over text generation in poem generation and question answering. \citet{han2022meet} focused on mimicking fictional characters through pseudo dialog prompting in capturing character styles. Similarly, \citet{madotto-etal-2021-continual} addressed continual learning in task-oriented dialogue systems with a benchmark featuring 37 domains and an efficient architectural method. \citet{yang-etal-2023-tailor} presents a parameter-efficient approach for attribute-controlled generation, achieving improvements in attribute-specific text generation with minimal additional training parameters.

\subsection{Multi-Agent Systems}
Our approach draws inspiration from relevant literature in the fields of multi-agent debating and simulation framework. \citet{chen2023agentverse} proposes a framework that demonstrates how multi-agent systems refine chatbot dynamics, promoting adaptive responses. \citet{shao-etal-2023-character} reveals the capability of fine-tuning models to emulate distinct personas, enriching the diversity of character simulations. Additionally, \citet{chan2023chateval} and \citet{du2023improving} highlight the role of these systems in improving evaluative and reasoning capacities of models through structured debates.  
Recent findings suggest that the performance of LLMs can be significantly enhanced by increasing the number of agents~\cite{li2024agents}. Through a simple sampling-and-voting method, it has been demonstrated that the efficacy of LLMs scales with the number of instantiated agents, an approach that is orthogonal to existing methods and suggests promising avenues for future research. In contrast to the common applications of multi-agent systems in evaluation and reasoning tasks, our research extends their utility to the mental health domain, addressing the intricate nature of psychological support with a richer, more context-sensitive dialogue framework.

\section{MentalAgora Framework}
\label{3-Method}
To enhance the capability of LLMs to generate tailored support responses, we propose~\mcraft, a framework simulating debating sessions with LLMs imitating strategic counselors. Since the structure of~\mcraft ~can consider diverse perspectives, it provides more contextually relevant and personalized support responses.
Our framework operates through three stages: \textbf{strategic debating}, \textbf{tailored counselor creation}, and \textbf{support response generation}. 

\subsection{Strategic Agents Setup}
We select counseling strategies \citep{Fall2022Theoretical,Ellis1989Rational} and compose attributes based on the strategies to ensure that each response is not only individually dynamic but also therapeutically appropriate.
There are three attributes used: reframing, which involves altering perspectives for a more positive outlook; unconditional positive regard, emphasizing total acceptance and support of the user; and solution-focused approaches, which concentrate on the user’s strengths and goals~\citep{zhou-etal-2023-facilitating,biyani-etal-2014-identifying}.
Each of these attributes is chosen for its proven efficacy in traditional counseling settings~\citep{Fall2022Theoretical,Ellis1989Rational}.
\cref{sec:90-Counseling-Attributes} provides detailed description for the attributes.

Building on these attributes, our counselor agents function based on their own attributes within a multi-agent role setup, utilizing prompting to incorporate core counseling strategies to analyze and address the concerns and needs of users.

\subsection{Stage 1: Strategic Debating}
In the strategic debating stage (\cref{alg:StrategicDebating}), an LLM functions as an independent agent representing each attribute through unique role settings reflecting the specific attribute.
Each agent is instructed to evaluate the user's status and concerns from its own perspective.
The debate is also structured to determine the relative importance of each attribute in addressing the user's concerns.
Each agent $a_j$ generates a response based on the user's post $P$ and the debating history $D$.
This process is repeated by the number of debating turns $N$.
Through this process, the debating history regarding the user post is obtained.

\begin{algorithm}[ht!]
\caption{Strategic Debating Stage}
\label{alg:StrategicDebating}
\begin{algorithmic}[1]
    \REQUIRE User post $P$, Number of debating turns $N$, Counselor agents $A = \{a_1, a_2, a_3\}$
    \ENSURE Debating dialogue history $D$
    \STATE $D = []$ \COMMENT{Initialize dialogue history list}  
    \FOR{$i$ from 1 to $N$}
        \FOR{each agent $a_j$ in $A$}  
            \STATE Generate response $R_{i, a_j}$ for $P$ with $a_j$ referring to $D$ 
            \STATE Add response $R_{i, a_j}$ to dialogue $D$  
        \ENDFOR  
    \ENDFOR
    \RETURN $D$
\end{algorithmic}
\end{algorithm}
\subsection{Stage 2: Tailored Counselor Creation}

\begin{figure}[ht!]
    \centering
    \includegraphics[width=1\linewidth]{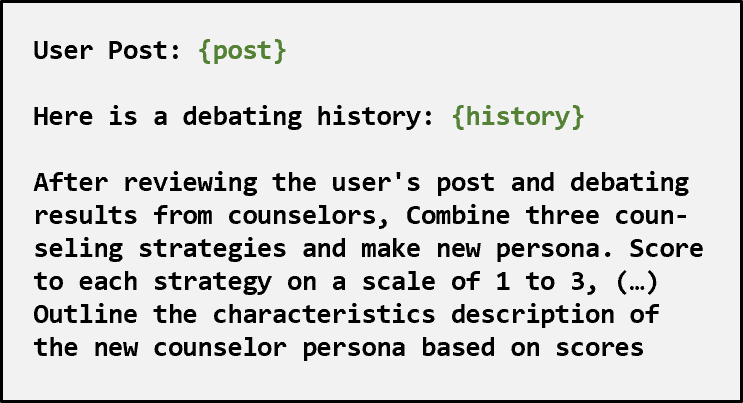}
    \caption{Prompt example for tailored counselor creation.
    The green texts are placeholders for actual data.
    }
    \label{fig:stage2-promp}
\end{figure}

This stage involves creating a prompt for a tailored counselor to generate actual user responses.
The LLM is instructed to reference the user's posts and debating history to determine the degree of influence of each agent and to generate the counselor prompt.
The degree of influence is expressed as a score between 1 and 3.
\cref{fig:stage2-promp} manifests our input prompt for this stage.

\subsection{Stage 3: Support Response Generation}

\begin{figure}[ht!]
    \centering
    \includegraphics[width=1\linewidth]{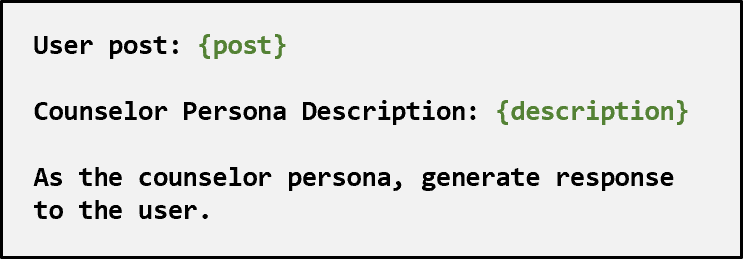}
    \caption{
    Prompt example for response generation.
    The green texts are placeholders for actual data.
    }
    \label{fig:stage3-promp}
\end{figure}

In the final step, we instruct the LLM to write a response that provides the most appropriate support for the user's post, with the prompt shown in \cref{fig:stage3-promp}. At this time, the LLM is given the previously created counselor prompt as a persona.
Since the prompt includes the influence scores for each attribute, the LLM can numerically reflect the importance of each attribute and also utilize the extensive user information embedded in debating history.

\section{Datasets}
\label{4-Dataset}
It is necessary for user posts to contain their personal mental concerns along with their individual characteristic information.
Additionally, for evaluating responses generated by the model, golden responses are required. These responses need to be tailored to each post, provide emotional empathy, and be clinically appropriate.
First, we used the Counsel Chat dataset \citep{bertagnolli2020counsel}, which consists of user concerns and responses from mental health experts on an online mental health platform.

We also build our own dataset \textbf{TherapyTalk}.
TherapyTalk consists of subreddit posts related to mental health concerns from Reddit and responses to these posts by mental health experts.
Unlike Counsel Chat, TherapyTalk is additionally labeled by annotators how much each response includes the three attributes. These attribute levels are used in our analysis to demonstrate similarity between our approach and expert responses.
We guide the experts to write consistent responses with guidelines.
\cref{91-Dataset-Creation} describes the dataset construction process in detail.

\begin{table*}[t!]
\centering
\resizebox{\linewidth}{!}{
\begin{tabular}{ll||ccccc||ccccc}
\toprule
\multicolumn{2}{l||}{ } & \multicolumn{5}{c||}{\textbf{TherapyTalk}} & \multicolumn{5}{c}{\textbf{Counsel Chat Dataset}} \\
\midrule
Model & Method & \multicolumn{1}{c}{BLEU} & \multicolumn{1}{c}{R-L} & \multicolumn{1}{c}{BScore} & \multicolumn{1}{c}{GM} & \multicolumn{1}{c||}{HM} & \multicolumn{1}{c}{BLEU} & \multicolumn{1}{c}{R-L} & \multicolumn{1}{c}{BScore} & \multicolumn{1}{c}{GM} & \multicolumn{1}{c}{HM} \\
\midrule
GPT-4.0 & SA & 24.52 & 15.86 & \underline{94.79} & 33.28 & 26.23 & 18.35 & 14.43 & 94.16 & 32.58 & 23.69 \\
& SAA & 24.23 & \textbf{16.51} & 94.71 & 33.59 & 26.69 &  18.05 & 14.47 & 94.20 & 32.45 & 23.62  \\
& MAA & \underline{25.27} & 16.94 & 94.70 & \underline{34.35} & \underline{27.48} & \underline{19.48} & \underline{15.36} & \textbf{95.32}  & \textbf{34.45} & \underline{25.42} \\
& \mcraft & \textbf{28.59} & \underline{16.50} & \textbf{95.31} & \textbf{35.56} & \textbf{28.28} & \textbf{19.53} & \textbf{15.40} & \underline{95.14}  & \textbf{34.45} & \textbf{25.45} \\
\midrule

GPT-3.5-turbo & SA & 21.18 & 15.09 & \textbf{94.83} & 31.18 & 24.19 & 15.99 & 14.08 & 94.04  & 29.18 & 20.63 \\
& SAA & 21.38 & \underline{15.34} & 94.45 & 31.41 & 24.48 & 15.04 & 13.77 & 94.14  & 28.55 & 19.83 \\
& MAA & \underline{22.95} & 15.25 & 94.77 & \underline{32.13} & \underline{25.06} & \underline{18.83} & \underline{14.30} & \underline{94.19}  & \underline{31.65} & \underline{22.78} \\
& \mcraft & \textbf{26.50} & \textbf{15.73} & \underline{94.80} & \textbf{34.06} & \textbf{26.82} & \textbf{19.14} & \textbf{14.87} & \textbf{94.22}  & \textbf{32.00} & \textbf{23.10} \\
\midrule

LLAMA-2-13b & SA & \underline{21.11} & 15.36 & 93.25 & \underline{31.15} & \underline{24.35} & 10.98 & 13.95 & 93.99 & 28.28 & 19.50  \\
& SAA & 10.52 & 15.01 & 94.08 & 24.58 & 17.41 & 08.76 & 12.80 & 94.06 & 26.22 & 15.91  \\
& MAA & 18.70 & \underline{15.52} & \underline{94.59} & 30.17 & 23.35 & \underline{16.35} & \underline{14.27} & \underline{94.15} & \underline{31.17}  & \underline{21.92} \\
& \mcraft & \textbf{26.49} & \textbf{15.65} & \textbf{94.61} & \textbf{33.98} & \textbf{26.73} & \textbf{17.74} & \textbf{14.29} & \textbf{94.21}  & \textbf{32.00} & \textbf{23.00} \\
\midrule

MentalAlpaca & SA & 18.42 & 15.33 & 94.96 & 29.93 & 23.07 & 10.74 & \underline{14.36} & 93.22  & 27.89 & 19.29\\
& SAA & 08.49 & 13.65 & \textbf{95.12} & 22.26 & 14.88 & 09.28 & 12.52 & 93.29 & 25.76  & 16.88 \\
& MAA & \underline{18.70} & \underline{15.52} & 94.59 & \underline{30.17} & \underline{23.35} & \underline{12.36} & 13.43 & \underline{94.18} & \underline{29.58} & \underline{20.08}  \\
& \mcraft & \textbf{20.65} & \textbf{15.90} & \underline{94.97} & \textbf{31.48} & \textbf{24.62} & \textbf{18.89} & \textbf{14.47} & \textbf{94.44}  & \textbf{32.54} & \textbf{23.54} \\
\bottomrule
\end{tabular}
}
\caption{
Comparison of various LLM configurations across four model settings on two counseling datasets using BLEU, ROUGE-L (R-L), and BERTScore (BScore), which measure the quality of the generated responses compared to the ground truth. GM and HM represent geometric mean and harmonic mean, respectively, of the evaluation metrics. \mcraft ~outperforms the other configurations for almost all metrics.
}
\label{tab:exp1-results}
\end{table*}

\begin{table}[t!]
    \centering
    \resizebox{\columnwidth}{!}{
\begin{tabular}{llllll|c}
\toprule
Method & Cus. & Sat. & Pro. & Rel. & Und. & Rank  \\ \midrule
SA & 3.72  & 3.74   & 3.51    & 3.41 & 3.38 & 2.68\\
SAA & 3.90   &  3.76    & 3.46    & 3.71$^\dagger$ & 3.33 &2.74 \\
MAA   & 3.92   & 3.96$^\dagger$    & 3.87$^\dagger$     & 3.68$^\dagger$ & 3.65 & 2.49  \\
\mcraft{}   & \textbf{4.03}    & \textbf{3.99}  & \textbf{3.99}     & \textbf{3.82} & \textbf{3.81} & \textbf{1.80}\\
\bottomrule
\end{tabular}
}
\caption{Human evaluation results on Customization (Cus.), Satisfaction (Sat.), Professionalism (Pro.), Relevance (Rel.), and Understanding (Und.), scored on a scale of 1 to 5, along with their overall preference ranks. All differences between our framework and the baselines are statistically significant with a $p$-value less than 0.05, except for those indicated with $\dagger$.}
\label{tab:exp-human}
\end{table}

\section{Experiments}
\label{5-Experiments}
We conduct two experiments to evaluate \mcraft, focusing on its effectiveness and controllability. We measure the framework's capacity to tailor responses that meet professional mental health care criteria and align with individual user preferences. Also, we examine our framework's controllability that indicating how well it reflects each attribute score provided as input into responses.

\paragraph{Models} To investigate the robustness and stability of \mcraft, we extend our research to include tests across various LLMs. The models are included GPT-3.5-turbo~\citep{ouyang2022training}, GPT-4.0~\citep{openai2024gpt4}, LLaMA-2-13B~\citep{touvron2023llama} and Mentalalpaca~\citep{Xu_2024}. Each of these models represents different architectures and capacities, providing a comprehensive basis for evaluating the framework's performance. 

\subsection{Tailored Support Response Generation}
\label{5-Experiments-1}

The investigation aims to validate the efficacy of our framework in providing mental health support grounded in expert principles and tailored to individual user needs.
The baseline settings used for the comparison are as follows.

\begin{itemize}
\setlength\itemsep{-0.1em}
\item \textit{Single Agent (SA)} generates responses without any specific attributes or iterative debating, serving as our baseline for fundamental response generation capabilities.

\item \textit{Single Agent Attributes (SAA)} enhances the single agent setup by incorporating attributes into the response generation, enabling us to compare the adaptability of a single agent in terms of ensuring customized response.

\item \textit{Multi Agents Attributes (MAA)} employs a multi-agent system but excludes the iterative debating component. This setup aims to assess the impact of debating history in our framework.
\end{itemize}

\paragraph{Automatic Evaluation}
\cref{tab:exp1-results} shows the automatic evaluation results. We utilize two benchmark datasets: TherapyTalk and a counseling utterance generation dataset~\citep{bertagnolli2020counsel}.
We evaluated the expert alignment of the methods by comparing the annotated responses of experts in the dataset across various metrics.
Evaluations using metrics such as Unigram BLEU~\citep{bleu}, ROUGE-L~\citep{rouge}, BERTScore~\cite{bert-score}, Geometric mean (GM), and Harmonic mean (HM) show that \mcraft{} achieves superior performance across the board, enhancing the alignment and relevance of responses. 

Regarding the impact of configurations, SAA improves the performance compared to SA setup. 
Also, \mcraft{} leads to better performance than MAA, supporting the value of interactive debate elements in response quality. LLaMA-2-13B and MentalAlpaca show variability in performance, with notable declines in SAA configuration. This may indicate difficulties in attribute integration by a single agent.

\paragraph{Human Evaluation}
We also evaluate the generated responses by three human judgments.
This evaluation involves a specialized form with guidelines and criteria to assess three baselines and \mcraft~responses. Evaluators score the responses on 5 likert scale based on following criteria detailed in \cref{92-Human-Evaluation}.
For the ranking task, evaluators are asked to order the responses from most to least effective based on their overall attractiveness and effectiveness assessment, identifying which response is most directly benefit and best-satisfied user preferences.

\cref{tab:exp-human} indicates the results of human evaluation. 
\mcraft{} excels in customizing responses across various dimensions.
It scores the highest in all the criteria, confirming \mcraft{}'s ability to adapt responses to individual user needs and maintain a high level of contextually appropriate engagement considering users' states.
The significant lead in the overall ranking highlights the framework's effectiveness and an advantage over the single agent setups or non-debating setups.
The inter-annotator reliability, assessed using Fleiss' kappa \citep{Fleiss1971}, is 0.558 overall.

\paragraph{LLM as a judge evaluation}
In addition, we employ an LLM as a judge approach~\citep{zheng2023judging}. This method involves using GPT-4.0-turbo to assess the generated responses using a 5-point Likert scale, enabling us to cross-validate the human judgments with automatic evaluations. \cref{tab:gpteval} shows the results that \mcraft{} outperforms other methods in all aspects.

\subsection{Attribute Controllable Generation}
\label{5-Experiments-2}
In \cref{5-Experiments-1}, we verify our framework's effectiveness in aligning with expert responses. Building on this, we conduct two further investigations. The first assesses whether our framework maintains this expert alignment even in terms of attribute scoring, thereby supporting its robustness in expert alignment.
The second experiment, a control test, checks if these attribute scores are correctly represented in the responses.
Through these two steps, we verify precise attribute classification and effective attribute exploitation skills, respectively.
Unlike \cref{5-Experiments-1}, configuration SA is omitted because it does not employ attributes.

\paragraph{Automatic Evaluation} 
We utilize MentalRoBERTa \cite{ji-etal-2022-mentalbert} fine-tuned on TherapyTalk dataset to predict attribute score of responses.
To assess the controllability, we estimate scores of responses from each method and human experts, enabling to compare attribute levels in responses.

\begin{figure}[t!]
    \centering
    \includegraphics[width=1\linewidth]{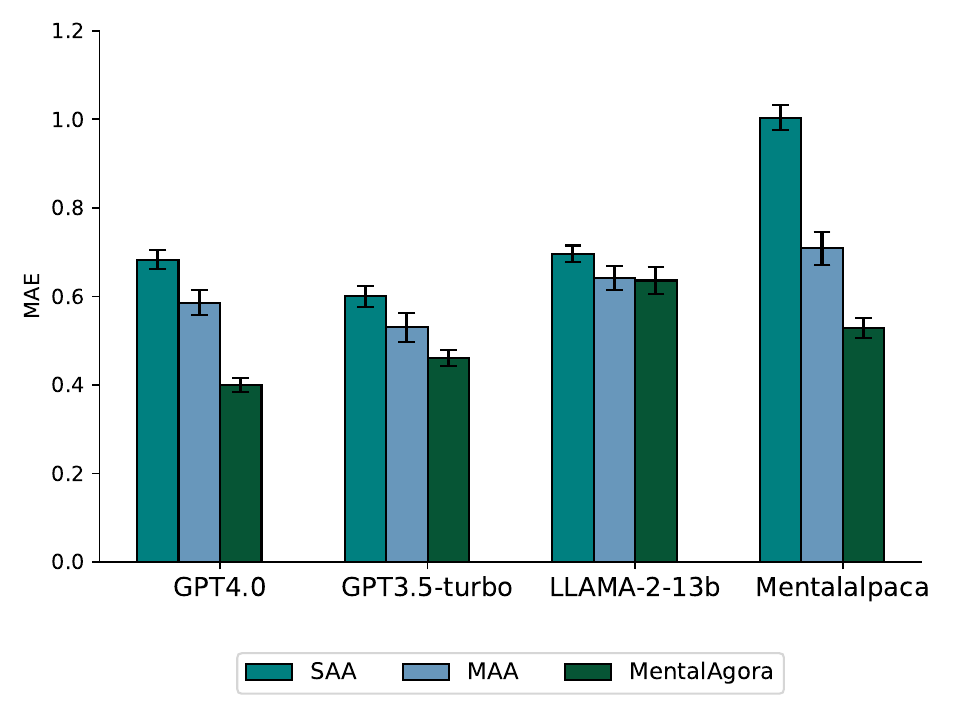}
    \caption{
    Automatic evaluation results of overall attribute control using Mean Absolute Error (MAE) which demonstrates the differences between scores from expert responses in TherapyTalk and generated responses across various LLMs and methods. MentalAgora outperforms all other configurations. 
    }
    \label{fig:ex2-maeplot-expert}
\end{figure}

\begin{figure}[t!]
    \centering
    \includegraphics[width=1\linewidth]{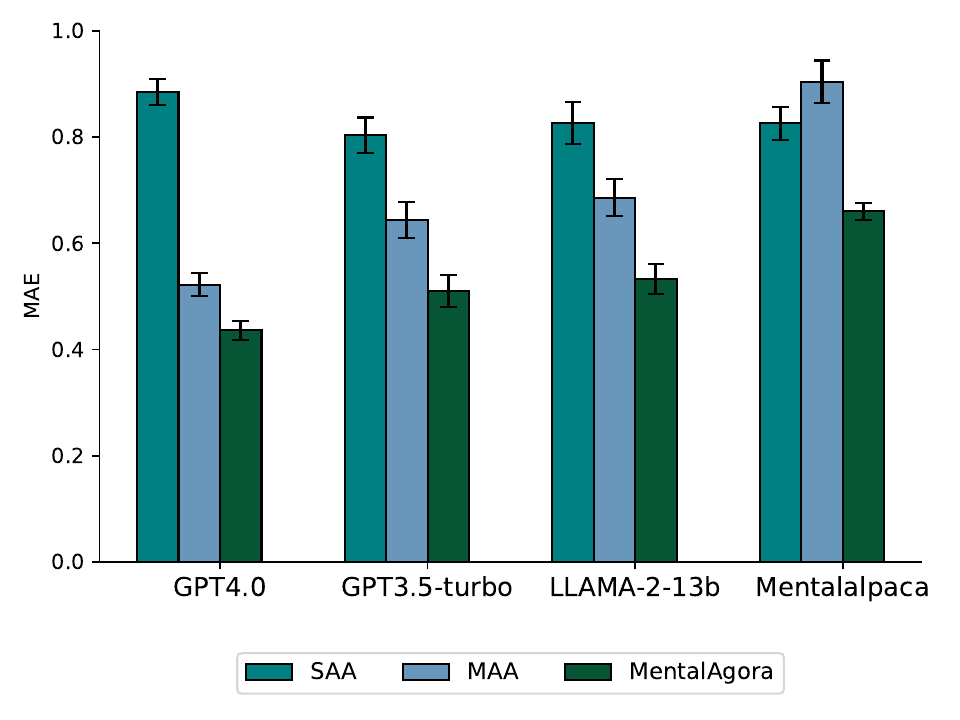}
    \caption{
    Automatic evaluation results of overall attribute control across LLMs and methods using MAE which measures the difference of attribute scores between generated responses and given input scores.
    \mcraft ~outperforms the other configurations compared.
    }
    \label{fig:ex2-maeplot}
\end{figure}

In \cref{fig:ex2-maeplot-expert}, the attribute levels of the methods are compared to ones from human experts by calculating MAE.
\mcraft{} shows the best performance, indicating its superior attribute integration capabilities. This suggests that \mcraft{} is effective in matching the attribute scores of expert responses, reflecting a high level of precision in response generation.

\cref{fig:ex2-maeplot} examines models' internal controllability of how much they can provide responses matching the input attribute scores in tailored counselor prompts.
MAE is difference between model input attribute scores and predicted scores of their output response.
The majority of methods exhibit higher MAE values, highlighting challenges in attribute control.
\mcraft{} stands out in this evaluation, which shows its effectiveness in reflecting the input scores into the generated responses.

\section{Analysis}
\label{sec:6-Analysis}
We conduct an ablation study focusing on how the inclusion and exclusion of specific attribute agents affect the generated responses in terms of controllability and interpretability. 
For a more in-depth exploration of the experimental results, \cref{tab:case_study} provides a comparative analysis influence of the inclusion or removal of specific attributes on responses. 

\subsection{Ablation Experimental Setup}
These are additional experimental setups implemented to further assess the controllability and interpretability of attribute integration. 
\begin{itemize}
\setlength\itemsep{-0.3em}

    \item \textit{Removal Attributes Setup} removes one attribute at a time for all possible combinations of two attributes to evaluate how the absence of each influences overall response performance and expert alignment.
    \item \textit{Uniform Attribute Setup} configures all agents to utilize the same attribute, allowing us to analyze how dominant the selected attribute becomes in the response generation process.
\end{itemize}

By adjusting the attributes in these specified setups, we demonstrate the framework’s ability to control and reflect these attributes in its outputs.

\begin{table}[t!]
    \centering
    \resizebox{0.8\columnwidth}{!}{
\begin{tabular}{lccc}
\toprule
Method & BLEU & R-L & BScore   \\ \midrule
\mcraft   & \textbf{28.59} & \textbf{16.50} & \textbf{95.31}\\
 - Reframing & 26.46 & 15.09 & 94.73 \\
 - Solution & 26.62 & 15.28 & 94.74 \\
 - Regard & 25.91 & 14.57 & 94.71 \\
Reframing only & 25.78 & 14.76 & 94.68 \\
Solution only & 25.30 & 14.37 & 94.72 \\
Regard only & 26.50 & 15.19 & 94.72 \\
\bottomrule
\end{tabular}
}
\caption{
Ablation study to evaluate the impact of each attribute by removing one and observing changes in task performance and alignment with desired outcomes. \mcraft{}, which combines three attributes, outperforms other methods.
}
\label{tab:ab-1}
\end{table}

\begin{table}[t!]
\centering
\resizebox{\linewidth}{!}{
\begin{tabular}{l|rrr|c}
\toprule
\multicolumn{1}{l|}{Method} & \multicolumn{1}{c}{Reframing} & \multicolumn{1}{c}{Solution} & \multicolumn{1}{c|}{Regard} & Total Diff.\\
\midrule
\mcraft           & -0.01 & -0.03 & +0.02 & \textbf{0.06} \\
 - Reframing      & \underline{-0.17} & +0.20 & -0.02 & 0.39 \\
 - Solution       & +0.10 & \underline{-0.33} & +0.09 & 0.52 \\
 - Regard         & +0.29 & +0.26 & \underline{-0.51} & 1.06 \\
Reframing only    & \underline{+0.63} & -0.14 & +0.05 & 0.69 \\
Solution only     & -0.31 & \underline{+0.40} & -0.03 & 0.74 \\
Regard only       & -0.08 & +0.10 & \underline{+0.10} & 0.28 \\
\bottomrule
\end{tabular}
}
\caption{
Comparison of methods with human experts' scores. Values represent the difference from human scores.
\mcraft{}~shows higher alignment with human scores than the other methods.
}
\label{tab:ab-2}
\end{table}

\subsection{Results and Analysis}
\paragraph{Expert-alignment Performance}

The performance evaluation in \cref{tab:ab-1} employs the same metrics and dataset as outlined in \cref{tab:exp1-results}.
For overall performance, utilizing all attributes in \mcraft{} configuration yields the highest performance metrics, indicating optimal framework operation. In contrast, selective removal of attributes decreases performance metrics, suggesting that balanced attribute configurations are crucial for maintaining performance. 

\paragraph{Attribute Controllability}

\cref{tab:ab-2} utilized expert response attribute scores from TherapyTalk dataset. The deltas calculated represent the mean of score differences between expert responses and those generated by the framework with ablation settings, with the total difference computed as the absolute sum of these deltas.
We observe that removing a specific attribute results in a divergence from human alignment and incorporating these attributes helps in aligning closer to expert-like responses.
Also, focusing on a single attribute fails to align with human responses, resulting in lower overall scores. 

Additionally, we observe in our experiment that when only one attribute is used, the corresponding score increases, and when a specific attribute is excluded, the score decreases. In either case, these changes lead to a divergence from human alignment.
This observation suggests that the framework's effective integration of various counseling strategies, represented as attributes in response generation, aligns well with expert views, supporting the notion that a diverse mix of counseling strategies enhances the comprehensiveness of handling multiple aspects of a user's condition~\cite{corey2016,norcross2005}.

\section{User Study}
\label{sec:7-user-study}
We verify the effectiveness of our responses through the datasets in various ways, but we cannot determine if the users who wrote the posts are really satisfied more than other baselines. We conduct a user study to assess if our method is effective in addressing the concerns of real people with actual problems.
We engage 18 participants in this study. Each participant is asked to provide a personal concern in a detailed format. 
Plus, we ask users to select which of the three attributes they most expected to be included in the response.

\begin{figure}[t!]
    \centering
    \includegraphics[width=1\linewidth]{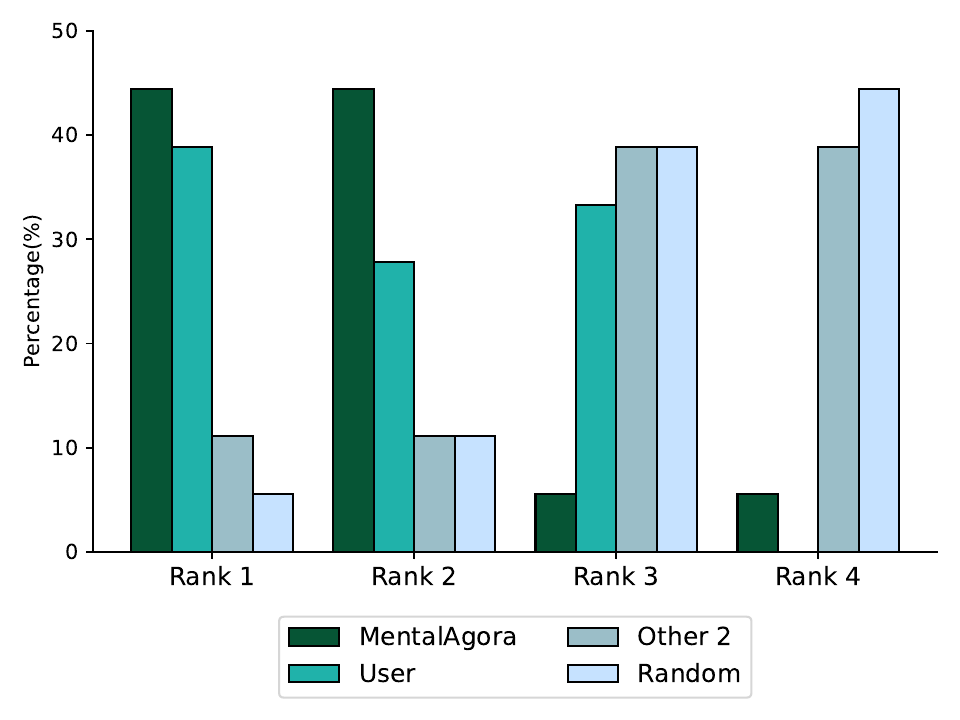}
    \caption{
    User study results show the preference rankings for different configurations: \mcraft{}, User Attribute, Other 2 attributes, and Random Attribute. The x-axis represents the rankings (1 to 4) while the y-axis shows the percentage of participants who ranked the method at that position.
    Users prefer \mcraft{} over the other baselines.
    }
    \label{fig:user-study}
\end{figure}

\subsection{User Study Setup}
Participants receive four types of responses to their concerns, each generated by differently configured versions of \mcraft{}. 
\begin{itemize}
\setlength\itemsep{-0.3em}
    \item \textit{User Attribute} generates responses based on a single attribute chosen by the user to observe user preference for responses aligned with their direct choices.
    \item \textit{Other 2 Attributes} use two attributes not selected by the user to see the preference changes when the user-selected attribute is excluded.
    \item \textit{Random Attribute} employs a randomly selected attribute except the attribute chosen by the user to assess the impact of responses that diverge from the user's preferences.
\end{itemize}

\subsection{Results and Analysis}

\cref{fig:user-study} demonstrates the user study results highlighting a clear preference trend towards \mcraft, which ranked as the most favored across all choices. 

\paragraph{User Preference}

The preference rankings favor \mcraft{}, which consistently secures the top position.
This reflects a high level of user preference for responses that integrate all attributes, suggesting that The response from \mcraft{} is not only well aligned with experts but also consistent with the user's preferences. The results illustrate the impact of attribute configuration on user preferences.
While \mcraft{} is most preferred, responses based on user attribute setup show moderate preferences.
Random and other 2 attributes are notably less preferred, emphasizing the importance of employing user-relevant attributes in achieving user satisfaction.

\paragraph{Variability in Preference}

We note that incorporating user attribute configurations results in a reduction in variance among the participant preferences. This finding indicates that knowing user preferences allows responses that align closely with user expectations and exhibit more consistency in satisfying diverse user needs. However, while \mcraft{} tends to rank highly in terms of user preference, it is also associated with a slightly higher variance. This suggests that \mcraft{} frequently delivers highly preferred responses, but the degree of satisfaction can vary among different users. This variability highlights that \mcraft{}, effectively meets the preferences of many users without direct user preference input ~\cite{gilbert2011}.

\section{Conclusion and Future Work}
\label{8-Conclusion}
Our framework generates expert-aligned and user preference-enhanced responses through a multi-agent system. Our experiments confirm the framework's capability to integrate complex user needs and maintain robust performance across different LLMs, showcasing its potential for real-world applications. Moving forward, we aim to enhance \mcraft{} by incorporating a wider range of attributes to increase diversity. Additionally, we plan to validate the framework's effectiveness in therapy by deploying it in actual clinical environments and assessing its impact on therapeutic outcomes.

\section*{Limitations}
\label{90-limitations}
Our study currently utilizes three distinct attributes based on well-known counseling methodologies. However, the use of counseling techniques is highly variable, dependent on individual preferences and specific issues~\cite{article-2,Swift2018}. This diversity necessitates the expansion of our framework to include a wider range of attributes to better cater to the varied needs of users. 
Additionally, we conduct our experiments with extensive text inputs that represent complex personal concerns, providing comprehensive answers that cover various aspects of the issues presented. This method effectively allows for a deep analysis of response capabilities within single posts. However, it lacks the capability to capture the dynamics typical of multi-turn interactions between a user and a virtual counselor. Future research should look to include interactive settings that facilitate real-time exchanges to enhance the realism and effectiveness of our model. Our dataset, which is composed of sets of three posts per virtual user, is designed to support such expansions, enabling more detailed studies on interactive counseling scenarios.

\section*{Ethics Statement}
\label{95-Ethics-statement}
\paragraph{\textbf{Data Use and Confidentiality}} Our research creates and uses the TherapyTalk dataset, which comprises anonymized posts containing personal information. We strictly prohibit using this data to identify individuals or for any purpose outside the intended scope of research. 
All examples of user posts used in this paper are rephrased.
Compliance with licensing agreements governing the use of this dataset is mandatory.

\paragraph{\textbf{IRB Approval}} The primary objective of our research is to mitigate emotional stress and offer potential solutions for those with mental health concerns. It is crucial to understand that our methods do not replace professional medical diagnosis or treatment. The research includes a user study that has been reviewed and approved by the Institutional Review Board (IRB), ensuring that all procedures adhere to the ethical standards required for human subject research.\footnote{IRB approval number: SKKU 2024-05-027}

\paragraph{\textbf{Ethical Use}} This study employs techniques to direct the responses of LLMs by adjusting attributes. These methods are designed to improve the relevance and supportiveness of responses within therapeutic contexts. It is critical to state that using these techniques to negatively affect individuals' mental health is strictly prohibited.

\bibliography{anthology,custom}

\begin{thebibliography}{55}
\providecommand{\natexlab}[1]{#1}

\bibitem[{Bannink(2007)}]{sfbt}
F.~P. Bannink. 2007.
\newblock \href {https://doi.org/10.1007/s10879-006-9040-y} {Solution-focused brief therapy.}
\newblock \emph{Journal of Contemporary Psychotherapy: On the Cutting Edge of Modern Developments in Psychotherapy}, 37(2):87--94.

\bibitem[{Beck(1976)}]{beck-ct}
Aaron~T. Beck. 1976.
\newblock \emph{Cognitive therapy and the emotional disorders.}
\newblock Cognitive therapy and the emotional disorders. International Universities Press, Oxford, England.

\bibitem[{Bertagnolli(2020)}]{bertagnolli2020counsel}
Nicolas Bertagnolli. 2020.
\newblock Counsel chat: Bootstrapping high-quality therapy data.

\bibitem[{Biyani et~al.(2014)Biyani, Caragea, Mitra, and Yen}]{biyani-etal-2014-identifying}
Prakhar Biyani, Cornelia Caragea, Prasenjit Mitra, and John Yen. 2014.
\newblock \href {https://aclanthology.org/C14-1079} {Identifying emotional and informational support in online health communities}.
\newblock In \emph{Proceedings of {COLING} 2014, the 25th International Conference on Computational Linguistics: Technical Papers}, pages 827--836, Dublin, Ireland. Dublin City University and Association for Computational Linguistics.

\bibitem[{Chan et~al.(2023)Chan, Chen, Su, Yu, Xue, Zhang, Fu, and Liu}]{chan2023chateval}
Chi-Min Chan, Weize Chen, Yusheng Su, Jianxuan Yu, Wei Xue, Shanghang Zhang, Jie Fu, and Zhiyuan Liu. 2023.
\newblock \href {https://arxiv.org/abs/2308.07201} {Chateval: Towards better llm-based evaluators through multi-agent debate}.
\newblock \emph{Preprint}, arXiv:2308.07201.

\bibitem[{Chen et~al.(2023{\natexlab{a}})Chen, Liu, Huang, Wu, Liu, Jiang, Pu, Lei, Chen, Wang, Lian, and Chen}]{chen2023large}
Jin Chen, Zheng Liu, Xu~Huang, Chenwang Wu, Qi~Liu, Gangwei Jiang, Yuanhao Pu, Yuxuan Lei, Xiaolong Chen, Xingmei Wang, Defu Lian, and Enhong Chen. 2023{\natexlab{a}}.
\newblock \href {https://arxiv.org/abs/2307.16376} {When large language models meet personalization: Perspectives of challenges and opportunities}.
\newblock \emph{Preprint}, arXiv:2307.16376.

\bibitem[{Chen et~al.(2023{\natexlab{b}})Chen, Su, Zuo, Yang, Yuan, Chan, Yu, Lu, Hung, Qian, Qin, Cong, Xie, Liu, Sun, and Zhou}]{chen2023agentverse}
Weize Chen, Yusheng Su, Jingwei Zuo, Cheng Yang, Chenfei Yuan, Chi-Min Chan, Heyang Yu, Yaxi Lu, Yi-Hsin Hung, Chen Qian, Yujia Qin, Xin Cong, Ruobing Xie, Zhiyuan Liu, Maosong Sun, and Jie Zhou. 2023{\natexlab{b}}.
\newblock \href {https://arxiv.org/abs/2308.10848} {Agentverse: Facilitating multi-agent collaboration and exploring emergent behaviors}.
\newblock \emph{Preprint}, arXiv:2308.10848.

\bibitem[{Chiu et~al.(2024)Chiu, Sharma, Lin, and Althoff}]{chiu2024computational}
Yu~Ying Chiu, Ashish Sharma, Inna~Wanyin Lin, and Tim Althoff. 2024.
\newblock \href {https://arxiv.org/abs/2401.00820} {A computational framework for behavioral assessment of llm therapists}.
\newblock \emph{Preprint}, arXiv:2401.00820.

\bibitem[{Corey(2016)}]{corey2016}
Gerald Corey. 2016.
\newblock \emph{Theory and Practice of Counseling and Psychotherapy}.
\newblock Cengage Learning, Boston.

\bibitem[{Du et~al.(2023)Du, Li, Torralba, Tenenbaum, and Mordatch}]{du2023improving}
Yilun Du, Shuang Li, Antonio Torralba, Joshua~B. Tenenbaum, and Igor Mordatch. 2023.
\newblock \href {https://arxiv.org/abs/2305.14325} {Improving factuality and reasoning in language models through multiagent debate}.
\newblock \emph{Preprint}, arXiv:2305.14325.

\bibitem[{Ellis(1989)}]{Ellis1989Rational}
Albert Ellis. 1989.
\newblock \href {https://doi.org/10.1007/BF01173455} {Rational-emotive therapy and cognitive behavior therapy: Similarities and differences}.
\newblock \emph{Cognitive Therapy and Research}, 13(5):325--340.

\bibitem[{Fall et~al.(2022)Fall, Holden, and Marquis}]{Fall2022Theoretical}
Kevin~A. Fall, Janice~Miner Holden, and Andre Marquis. 2022.
\newblock \href {https://www.routledge.com/Theoretical-Models-of-Counseling-and-Psychotherapy/Fall-Holden-Marquis/p/book/9781032038483} {\emph{Theoretical Models of Counseling and Psychotherapy}}, 4 edition.
\newblock Routledge.

\bibitem[{Fleiss(1971)}]{Fleiss1971}
Joseph~L. Fleiss. 1971.
\newblock \href {https://doi.org/10.1037/h0031619} {Measuring nominal scale agreement among many raters}.
\newblock \emph{Psychological Bulletin}, 76(5):378.

\bibitem[{Gilbert and Orlans(2011)}]{gilbert2011}
Maria Gilbert and Vanja Orlans. 2011.
\newblock \emph{Integrative Therapy: 100 Key Points and Techniques}.
\newblock Routledge, London.

\bibitem[{Guo et~al.(2024)Guo, Chen, Wang, Chang, Pei, Chawla, Wiest, and Zhang}]{guo2024large}
Taicheng Guo, Xiuying Chen, Yaqi Wang, Ruidi Chang, Shichao Pei, Nitesh~V. Chawla, Olaf Wiest, and Xiangliang Zhang. 2024.
\newblock \href {https://arxiv.org/abs/2402.01680} {Large language model based multi-agents: A survey of progress and challenges}.
\newblock \emph{Preprint}, arXiv:2402.01680.

\bibitem[{Han et~al.(2022)Han, Kim, Yoo, Seo, Kim, Erdenee, and Chang}]{han2022meet}
Seungju Han, Beomsu Kim, Jin~Yong Yoo, Seokjun Seo, Sangbum Kim, Enkhbayar Erdenee, and Buru Chang. 2022.
\newblock \href {https://arxiv.org/abs/2204.10825} {Meet your favorite character: Open-domain chatbot mimicking fictional characters with only a few utterances}.
\newblock \emph{Preprint}, arXiv:2204.10825.

\bibitem[{Hua et~al.(2024)Hua, Liu, Yang, Li, han Sheu, Zhou, Moran, Ananiadou, and Beam}]{hua2024large}
Yining Hua, Fenglin Liu, Kailai Yang, Zehan Li, Yi~han Sheu, Peilin Zhou, Lauren~V. Moran, Sophia Ananiadou, and Andrew Beam. 2024.
\newblock \href {https://arxiv.org/abs/2401.02984} {Large language models in mental health care: a scoping review}.
\newblock \emph{Preprint}, arXiv:2401.02984.

\bibitem[{Ji et~al.(2022{\natexlab{a}})Ji, Zhang, Ansari, Fu, Tiwari, and Cambria}]{ji-etal-2022-mentalbert}
Shaoxiong Ji, Tianlin Zhang, Luna Ansari, Jie Fu, Prayag Tiwari, and Erik Cambria. 2022{\natexlab{a}}.
\newblock \href {https://aclanthology.org/2022.lrec-1.778} {{M}ental{BERT}: Publicly available pretrained language models for mental healthcare}.
\newblock In \emph{Proceedings of the Thirteenth Language Resources and Evaluation Conference}, pages 7184--7190, Marseille, France. European Language Resources Association.

\bibitem[{Ji et~al.(2022{\natexlab{b}})Ji, Zhang, Ansari, Fu, Tiwari, and Cambria}]{mental-bert}
Shaoxiong Ji, Tianlin Zhang, Luna Ansari, Jie Fu, Prayag Tiwari, and Erik Cambria. 2022{\natexlab{b}}.
\newblock {MentalBERT: Publicly Available Pretrained Language Models for Mental Healthcare}.
\newblock In \emph{Proceedings of LREC}.

\bibitem[{Li et~al.(2023)Li, Chong, Stepputtis, Campbell, Hughes, Lewis, and Sycara}]{li-etal-2023-theory}
Huao Li, Yu~Chong, Simon Stepputtis, Joseph Campbell, Dana Hughes, Charles Lewis, and Katia Sycara. 2023.
\newblock \href {https://doi.org/10.18653/v1/2023.emnlp-main.13} {Theory of mind for multi-agent collaboration via large language models}.
\newblock In \emph{Proceedings of the 2023 Conference on Empirical Methods in Natural Language Processing}, pages 180--192, Singapore. Association for Computational Linguistics.

\bibitem[{Li et~al.(2024)Li, Zhang, Yu, Fu, and Ye}]{li2024agents}
Junyou Li, Qin Zhang, Yangbin Yu, Qiang Fu, and Deheng Ye. 2024.
\newblock \href {https://arxiv.org/abs/2402.05120} {More agents is all you need}.
\newblock \emph{Preprint}, arXiv:2402.05120.

\bibitem[{Lin(2004)}]{rouge}
Chin-Yew Lin. 2004.
\newblock \href {https://aclanthology.org/W04-1013} {{ROUGE}: A package for automatic evaluation of summaries}.
\newblock In \emph{Text Summarization Branches Out}, pages 74--81, Barcelona, Spain. Association for Computational Linguistics.

\bibitem[{Liu et~al.(2023)Liu, Li, Cao, Ren, Liao, and Wu}]{liu2023chatcounselor}
June~M. Liu, Donghao Li, He~Cao, Tianhe Ren, Zeyi Liao, and Jiamin Wu. 2023.
\newblock \href {https://arxiv.org/abs/2309.15461} {Chatcounselor: A large language models for mental health support}.
\newblock \emph{Preprint}, arXiv:2309.15461.

\bibitem[{Loh and Raamkumar(2023)}]{loh2023harnessing}
Siyuan~Brandon Loh and Aravind~Sesagiri Raamkumar. 2023.
\newblock \href {https://arxiv.org/abs/2310.08017} {Harnessing large language models' empathetic response generation capabilities for online mental health counselling support}.
\newblock \emph{Preprint}, arXiv:2310.08017.

\bibitem[{Low et~al.(2020)Low, Rumker, Torous, Cecchi, Ghosh, and Talkar}]{reddit-mental-health-dataset}
Daniel~M Low, Laurie Rumker, John Torous, Guillermo Cecchi, Satrajit~S Ghosh, and Tanya Talkar. 2020.
\newblock Natural language processing reveals vulnerable mental health support groups and heightened health anxiety on reddit during covid-19: Observational study.
\newblock \emph{Journal of medical Internet research}, 22(10):e22635.

\bibitem[{Lu et~al.(2023)Lu, Wei, Qu, Mao, Chen, and Chen}]{lu-etal-2023-miracle}
Zhenyi Lu, Wei Wei, Xiaoye Qu, Xian-Ling Mao, Dangyang Chen, and Jixiong Chen. 2023.
\newblock \href {https://doi.org/10.18653/v1/2023.findings-emnlp.395} {Miracle: Towards personalized dialogue generation with latent-space multiple personal attribute control}.
\newblock In \emph{Findings of the Association for Computational Linguistics: EMNLP 2023}, pages 5933--5957, Singapore. Association for Computational Linguistics.

\bibitem[{Madotto et~al.(2021)Madotto, Lin, Zhou, Moon, Crook, Liu, Yu, Cho, Fung, and Wang}]{madotto-etal-2021-continual}
Andrea Madotto, Zhaojiang Lin, Zhenpeng Zhou, Seungwhan Moon, Paul Crook, Bing Liu, Zhou Yu, Eunjoon Cho, Pascale Fung, and Zhiguang Wang. 2021.
\newblock \href {https://doi.org/10.18653/v1/2021.emnlp-main.590} {Continual learning in task-oriented dialogue systems}.
\newblock In \emph{Proceedings of the 2021 Conference on Empirical Methods in Natural Language Processing}, pages 7452--7467, Online and Punta Cana, Dominican Republic. Association for Computational Linguistics.

\bibitem[{McGrath et~al.(2023)McGrath, Al-Hamzawi, Alonso, Altwaijri, Andrade, Bromet, Bruffaerts, {de Almeida}, Chardoul, Chiu, Degenhardt, Demler, Ferry, Gureje, Haro, Karam, Karam, Khaled, Kovess-Masfety, Magno, Medina-Mora, Moskalewicz, Navarro-Mateu, Nishi, Plana-Ripoll, Posada-Villa, Rapsey, Sampson, Stagnaro, Stein, {ten Have}, Torres, Vladescu, Woodruff, Zarkov, Kessler, Aguilar-Gaxiola, Al-Hamzawi, Alonso, Altwaijri, Andrade, Atwoli, Benjet, Bromet, Bruffaerts, Bunting, de~Almeida, Cardoso, Chardoul, Cía, Degenhardt, {De Girolamo}, Gureje, Haro, Harris, Hinkov, yi~Hu, {De Jonge}, Karam, Karam, Karam, Kazdin, Kawakami, Kessler, Kiejna, Kovess-Masfety, McGrath, Medina-Mora, Moskalewicz, Navarro-Mateu, Nishi, Piazza, Posada-Villa, Scott, Stagnaro, Stein, {Ten Have}, Torres, Viana, Vigo, Vladescu, Williams, Woodruff, Wojtyniak, Xavier, and Zaslavsky}]{MCGRATH2023668}
John~J McGrath, Ali Al-Hamzawi, Jordi Alonso, Yasmin Altwaijri, Laura~H Andrade, Evelyn~J Bromet, Ronny Bruffaerts, José Miguel~Caldas {de Almeida}, Stephanie Chardoul, Wai~Tat Chiu, Louisa Degenhardt, Olga~V Demler, Finola Ferry, Oye Gureje, Josep~Maria Haro, Elie~G Karam, Georges Karam, Salma~M Khaled, Viviane Kovess-Masfety, Marta Magno, Maria~Elena Medina-Mora, Jacek Moskalewicz, Fernando Navarro-Mateu, Daisuke Nishi, Oleguer Plana-Ripoll, José Posada-Villa, Charlene Rapsey, Nancy~A Sampson, Juan~Carlos Stagnaro, Dan~J Stein, Margreet {ten Have}, Yolanda Torres, Cristian Vladescu, Peter~W Woodruff, Zahari Zarkov, Ronald~C Kessler, Sergio Aguilar-Gaxiola, Ali Al-Hamzawi, Jordi Alonso, Yasmin~A. Altwaijri, Laura~Helena Andrade, Lukoye Atwoli, Corina Benjet, Evelyn~J. Bromet, Ronny Bruffaerts, Brendan Bunting, José Miguel~Caldas de~Almeida, Graça Cardoso, Stephanie Chardoul, Alfredo~H. Cía, Louisa Degenhardt, Giovanni {De Girolamo}, Oye Gureje, Josep~Maria Haro, Meredith~G. Harris, Hristo Hinkov, Chi
  yi~Hu, Peter {De Jonge}, Aimee~N. Karam, Elie~G. Karam, Georges Karam, Alan~E. Kazdin, Norito Kawakami, Ronald~C. Kessler, Andrzej Kiejna, Viviane Kovess-Masfety, John~J. McGrath, Maria~Elena Medina-Mora, Jacek Moskalewicz, Fernando Navarro-Mateu, Daisuke Nishi, Marina Piazza, José Posada-Villa, Kate~M. Scott, Juan~Carlos Stagnaro, Dan~J. Stein, Margreet {Ten Have}, Yolanda Torres, Maria~Carmen Viana, Daniel~V. Vigo, Cristian Vladescu, David~R. Williams, Peter Woodruff, Bogdan Wojtyniak, Miguel Xavier, and Alan~M. Zaslavsky. 2023.
\newblock \href {https://doi.org/10.1016/S2215-0366(23)00193-1} {Age of onset and cumulative risk of mental disorders: a cross-national analysis of population surveys from 29 countries}.
\newblock \emph{The Lancet Psychiatry}, 10(9):668--681.

\bibitem[{Min et~al.(2022)Min, P{\'e}rez-Rosas, Resnicow, and Mihalcea}]{min-etal-2022-pair}
Do~June Min, Ver{\'o}nica P{\'e}rez-Rosas, Kenneth Resnicow, and Rada Mihalcea. 2022.
\newblock \href {https://doi.org/10.18653/v1/2022.emnlp-main.11} {{PAIR}: Prompt-aware marg{I}n ranking for counselor reflection scoring in motivational interviewing}.
\newblock In \emph{Proceedings of the 2022 Conference on Empirical Methods in Natural Language Processing}, pages 148--158, Abu Dhabi, United Arab Emirates. Association for Computational Linguistics.

\bibitem[{Mishra et~al.(2023)Mishra, Priya, and Ekbal}]{mishra-etal-2023-pal}
Kshitij Mishra, Priyanshu Priya, and Asif Ekbal. 2023.
\newblock \href {https://doi.org/10.18653/v1/2023.acl-long.685} {{PAL} to lend a helping hand: Towards building an emotion adaptive polite and empathetic counseling conversational agent}.
\newblock In \emph{Proceedings of the 61st Annual Meeting of the Association for Computational Linguistics (Volume 1: Long Papers)}, pages 12254--12271, Toronto, Canada. Association for Computational Linguistics.

\bibitem[{Norcross and Goldfried(2005)}]{norcross2005}
John~C. Norcross and Marvin~R. Goldfried. 2005.
\newblock \emph{Handbook of Psychotherapy Integration}.
\newblock Oxford University Press, New York.

\bibitem[{OpenAI(2024)}]{openai2024gpt4}
OpenAI. 2024.
\newblock \href {https://arxiv.org/abs/2303.08774} {Gpt-4 technical report}.
\newblock \emph{Preprint}, arXiv:2303.08774.

\bibitem[{Ouyang et~al.(2022)Ouyang, Wu, Jiang, Almeida, Wainwright, Mishkin, Zhang, Agarwal, Slama, Ray, Schulman, Hilton, Kelton, Miller, Simens, Askell, Welinder, Christiano, Leike, and Lowe}]{ouyang2022training}
Long Ouyang, Jeff Wu, Xu~Jiang, Diogo Almeida, Carroll~L. Wainwright, Pamela Mishkin, Chong Zhang, Sandhini Agarwal, Katarina Slama, Alex Ray, John Schulman, Jacob Hilton, Fraser Kelton, Luke Miller, Maddie Simens, Amanda Askell, Peter Welinder, Paul Christiano, Jan Leike, and Ryan Lowe. 2022.
\newblock \href {https://arxiv.org/abs/2203.02155} {Training language models to follow instructions with human feedback}.
\newblock \emph{Preprint}, arXiv:2203.02155.

\bibitem[{Papineni et~al.(2002)Papineni, Roukos, Ward, and Zhu}]{bleu}
Kishore Papineni, Salim Roukos, Todd Ward, and Wei-Jing Zhu. 2002.
\newblock \href {https://doi.org/10.3115/1073083.1073135} {Bleu: a method for automatic evaluation of machine translation}.
\newblock In \emph{Proceedings of the 40th Annual Meeting on Association for Computational Linguistics}, ACL '02, page 311–318, USA. Association for Computational Linguistics.

\bibitem[{Park et~al.(2023)Park, O'Brien, Cai, Morris, Liang, and Bernstein}]{park2023generative}
Joon~Sung Park, Joseph~C. O'Brien, Carrie~J. Cai, Meredith~Ringel Morris, Percy Liang, and Michael~S. Bernstein. 2023.
\newblock \href {https://arxiv.org/abs/2304.03442} {Generative agents: Interactive simulacra of human behavior}.
\newblock \emph{Preprint}, arXiv:2304.03442.

\bibitem[{Rashkin et~al.(2019)Rashkin, Smith, Li, and Boureau}]{rashkin2019empathetic}
Hannah Rashkin, Eric~Michael Smith, Margaret Li, and Y-Lan Boureau. 2019.
\newblock \href {https://arxiv.org/abs/1811.00207} {Towards empathetic open-domain conversation models: a new benchmark and dataset}.
\newblock \emph{Preprint}, arXiv:1811.00207.

\bibitem[{Raskin and Rogers(2005)}]{rogers-pct}
Nathaniel~J. Raskin and Carl~R. Rogers. 2005.
\newblock \href {https://doi.org/10.1177/009318530403200411} {\emph{Person-centered therapy.}}, pages 130--165.
\newblock Current psychotherapies, 7th ed., instr. ed. Thomson Brooks/Cole Publishing Co, Belmont, CA, US.

\bibitem[{Shao et~al.(2023)Shao, Li, Dai, and Qiu}]{shao-etal-2023-character}
Yunfan Shao, Linyang Li, Junqi Dai, and Xipeng Qiu. 2023.
\newblock \href {https://aclanthology.org/2023.emnlp-main.814} {Character-{LLM}: A trainable agent for role-playing}.
\newblock In \emph{Proceedings of the 2023 Conference on Empirical Methods in Natural Language Processing}, pages 13153--13187, Singapore. Association for Computational Linguistics.

\bibitem[{Sharma et~al.(2021)Sharma, Lin, Miner, Atkins, and Althoff}]{sharma2021facilitating}
Ashish Sharma, Inna~W. Lin, Adam~S. Miner, David~C. Atkins, and Tim Althoff. 2021.
\newblock \href {https://arxiv.org/abs/2101.07714} {Towards facilitating empathic conversations in online mental health support: A reinforcement learning approach}.
\newblock \emph{Preprint}, arXiv:2101.07714.

\bibitem[{Sharma et~al.(2023)Sharma, Rushton, Lin, Wadden, Lucas, Miner, Nguyen, and Althoff}]{sharma-etal-2023-cognitive}
Ashish Sharma, Kevin Rushton, Inna Lin, David Wadden, Khendra Lucas, Adam Miner, Theresa Nguyen, and Tim Althoff. 2023.
\newblock \href {https://doi.org/10.18653/v1/2023.acl-long.555} {Cognitive reframing of negative thoughts through human-language model interaction}.
\newblock In \emph{Proceedings of the 61st Annual Meeting of the Association for Computational Linguistics (Volume 1: Long Papers)}, pages 9977--10000, Toronto, Canada. Association for Computational Linguistics.

\bibitem[{Shen et~al.(2020)Shen, Welch, Mihalcea, and P{\'e}rez-Rosas}]{shen-etal-2020-counseling}
Siqi Shen, Charles Welch, Rada Mihalcea, and Ver{\'o}nica P{\'e}rez-Rosas. 2020.
\newblock \href {https://doi.org/10.18653/v1/2020.sigdial-1.2} {Counseling-style reflection generation using generative pretrained transformers with augmented context}.
\newblock In \emph{Proceedings of the 21th Annual Meeting of the Special Interest Group on Discourse and Dialogue}, pages 10--20, 1st virtual meeting. Association for Computational Linguistics.

\bibitem[{Swift et~al.(2018)Swift, Callahan, Cooper, and Parkin}]{Swift2018}
Joshua~K. Swift, Jennifer~L. Callahan, Michael Cooper, and Susannah Parkin. 2018.
\newblock \href {https://doi.org/10.1002/jclp.22680} {The impact of accommodating client preference in psychotherapy: A meta-analysis}.
\newblock \emph{Journal of Clinical Psychology}, 74(11):1925--1937.

\bibitem[{Tompkins et~al.(2013)Tompkins, Swift, and Callahan}]{article-2}
Kelley Tompkins, Joshua Swift, and Jennifer Callahan. 2013.
\newblock \href {https://doi.org/10.1037/a0032031} {Working with clients by incorporating their preferences}.
\newblock \emph{Psychotherapy (Chicago, Ill.)}, 50:279--83.

\bibitem[{Touvron et~al.(2023)Touvron, Martin, Stone, Albert, Almahairi, Babaei, Bashlykov, Batra, Bhargava, Bhosale, Bikel, Blecher, Ferrer, Chen, Cucurull, Esiobu, Fernandes, Fu, Fu, Fuller, Gao, Goswami, Goyal, Hartshorn, Hosseini, Hou, Inan, Kardas, Kerkez, Khabsa, Kloumann, Korenev, Koura, Lachaux, Lavril, Lee, Liskovich, Lu, Mao, Martinet, Mihaylov, Mishra, Molybog, Nie, Poulton, Reizenstein, Rungta, Saladi, Schelten, Silva, Smith, Subramanian, Tan, Tang, Taylor, Williams, Kuan, Xu, Yan, Zarov, Zhang, Fan, Kambadur, Narang, Rodriguez, Stojnic, Edunov, and Scialom}]{touvron2023llama}
Hugo Touvron, Louis Martin, Kevin Stone, Peter Albert, Amjad Almahairi, Yasmine Babaei, Nikolay Bashlykov, Soumya Batra, Prajjwal Bhargava, Shruti Bhosale, Dan Bikel, Lukas Blecher, Cristian~Canton Ferrer, Moya Chen, Guillem Cucurull, David Esiobu, Jude Fernandes, Jeremy Fu, Wenyin Fu, Brian Fuller, Cynthia Gao, Vedanuj Goswami, Naman Goyal, Anthony Hartshorn, Saghar Hosseini, Rui Hou, Hakan Inan, Marcin Kardas, Viktor Kerkez, Madian Khabsa, Isabel Kloumann, Artem Korenev, Punit~Singh Koura, Marie-Anne Lachaux, Thibaut Lavril, Jenya Lee, Diana Liskovich, Yinghai Lu, Yuning Mao, Xavier Martinet, Todor Mihaylov, Pushkar Mishra, Igor Molybog, Yixin Nie, Andrew Poulton, Jeremy Reizenstein, Rashi Rungta, Kalyan Saladi, Alan Schelten, Ruan Silva, Eric~Michael Smith, Ranjan Subramanian, Xiaoqing~Ellen Tan, Binh Tang, Ross Taylor, Adina Williams, Jian~Xiang Kuan, Puxin Xu, Zheng Yan, Iliyan Zarov, Yuchen Zhang, Angela Fan, Melanie Kambadur, Sharan Narang, Aurelien Rodriguez, Robert Stojnic, Sergey Edunov, and Thomas
  Scialom. 2023.
\newblock \href {https://arxiv.org/abs/2307.09288} {Llama 2: Open foundation and fine-tuned chat models}.
\newblock \emph{Preprint}, arXiv:2307.09288.

\bibitem[{Wang et~al.(2023)Wang, Wang, Mi, Deng, Wang, Liang, Xu, and Wong}]{wang-etal-2023-cue}
Hongru Wang, Rui Wang, Fei Mi, Yang Deng, Zezhong Wang, Bin Liang, Ruifeng Xu, and Kam-Fai Wong. 2023.
\newblock \href {https://doi.org/10.18653/v1/2023.findings-emnlp.806} {Cue-{C}o{T}: Chain-of-thought prompting for responding to in-depth dialogue questions with {LLM}s}.
\newblock In \emph{Findings of the Association for Computational Linguistics: EMNLP 2023}, pages 12047--12064, Singapore. Association for Computational Linguistics.

\bibitem[{Wei et~al.(2021)Wei, Liu, Mao, Guo, Zhu, Zhou, and Hu}]{wei2021emotionaware}
Wei Wei, Jiayi Liu, Xianling Mao, Guibing Guo, Feida Zhu, Pan Zhou, and Yuchong Hu. 2021.
\newblock \href {https://arxiv.org/abs/2106.03044} {Emotion-aware chat machine: Automatic emotional response generation for human-like emotional interaction}.
\newblock \emph{Preprint}, arXiv:2106.03044.

\bibitem[{WHO(2022)}]{who2022}
World Health~Organization WHO. 2022.
\newblock Mental health and covid-19: Early evidence of the pandemic’s impact.
\newblock https://iris.who.int/bitstream/handle/10665/352189/WHO-2019-nCoV-Sci-Brief-Mental-health-2022.1-eng.pdf?sequence=1.

\bibitem[{Xu et~al.(2024)Xu, Yao, Dong, Gabriel, Yu, Hendler, Ghassemi, Dey, and Wang}]{Xu_2024}
Xuhai Xu, Bingsheng Yao, Yuanzhe Dong, Saadia Gabriel, Hong Yu, James Hendler, Marzyeh Ghassemi, Anind~K. Dey, and Dakuo Wang. 2024.
\newblock \href {https://doi.org/10.1145/3643540} {Mental-llm: Leveraging large language models for mental health prediction via online text data}.
\newblock \emph{Proceedings of the ACM on Interactive, Mobile, Wearable and Ubiquitous Technologies}, 8(1):1–32.

\bibitem[{Yang et~al.(2023)Yang, Liu, Lei, Yang, Xue, Chen, and Xie}]{yang-etal-2023-tailor}
Kexin Yang, Dayiheng Liu, Wenqiang Lei, Baosong Yang, Mingfeng Xue, Boxing Chen, and Jun Xie. 2023.
\newblock \href {https://doi.org/10.18653/v1/2023.acl-long.25} {Tailor: A soft-prompt-based approach to attribute-based controlled text generation}.
\newblock In \emph{Proceedings of the 61st Annual Meeting of the Association for Computational Linguistics (Volume 1: Long Papers)}, pages 410--427, Toronto, Canada. Association for Computational Linguistics.

\bibitem[{Zhan et~al.(2022)Zhan, Sosea, Caragea, and Li}]{zhan2022feel}
Hongli Zhan, Tiberiu Sosea, Cornelia Caragea, and Junyi~Jessy Li. 2022.
\newblock \href {https://arxiv.org/abs/2210.12531} {Why do you feel this way? summarizing triggers of emotions in social media posts}.
\newblock \emph{Preprint}, arXiv:2210.12531.

\bibitem[{Zhang et~al.(2024)Zhang, Xu, Zhang, Liu, Hooi, and Deng}]{zhang2024exploring}
Jintian Zhang, Xin Xu, Ningyu Zhang, Ruibo Liu, Bryan Hooi, and Shumin Deng. 2024.
\newblock \href {https://arxiv.org/abs/2310.02124} {Exploring collaboration mechanisms for llm agents: A social psychology view}.
\newblock \emph{Preprint}, arXiv:2310.02124.

\bibitem[{Zhang* et~al.(2020)Zhang*, Kishore*, Wu*, Weinberger, and Artzi}]{bert-score}
Tianyi Zhang*, Varsha Kishore*, Felix Wu*, Kilian~Q. Weinberger, and Yoav Artzi. 2020.
\newblock \href {https://openreview.net/forum?id=SkeHuCVFDr} {Bertscore: Evaluating text generation with bert}.
\newblock In \emph{International Conference on Learning Representations}.

\bibitem[{Zheng et~al.(2023)Zheng, Chiang, Sheng, Zhuang, Wu, Zhuang, Lin, Li, Li, Xing, Zhang, Gonzalez, and Stoica}]{zheng2023judging}
Lianmin Zheng, Wei-Lin Chiang, Ying Sheng, Siyuan Zhuang, Zhanghao Wu, Yonghao Zhuang, Zi~Lin, Zhuohan Li, Dacheng Li, Eric~P. Xing, Hao Zhang, Joseph~E. Gonzalez, and Ion Stoica. 2023.
\newblock \href {https://arxiv.org/abs/2306.05685} {Judging llm-as-a-judge with mt-bench and chatbot arena}.
\newblock \emph{Preprint}, arXiv:2306.05685.

\bibitem[{Zhou et~al.(2023)Zhou, Chen, Wang, and Huang}]{zhou-etal-2023-facilitating}
Jinfeng Zhou, Zhuang Chen, Bo~Wang, and Minlie Huang. 2023.
\newblock \href {https://doi.org/10.18653/v1/2023.acl-long.96} {Facilitating multi-turn emotional support conversation with positive emotion elicitation: A reinforcement learning approach}.
\newblock In \emph{Proceedings of the 61st Annual Meeting of the Association for Computational Linguistics (Volume 1: Long Papers)}, pages 1714--1729, Toronto, Canada. Association for Computational Linguistics.

\bibitem[{Zou et~al.(2021)Zou, Yin, Zhong, Yang, Yang, and Tang}]{Zou_2021}
Xu~Zou, Da~Yin, Qingyang Zhong, Hongxia Yang, Zhilin Yang, and Jie Tang. 2021.
\newblock \href {https://doi.org/10.1145/3447548.3467418} {Controllable generation from pre-trained language models via inverse prompting}.
\newblock In \emph{Proceedings of the 27th ACM SIGKDD Conference on Knowledge Discovery \& Data Mining}, KDD '21, page 2450–2460, New York, NY, USA. Association for Computing Machinery.

\end{thebibliography}
\appendix

\newpage

\section{Counseling Attributes}
\label{sec:90-Counseling-Attributes}
We have choose three attributes to differentiate each LLM agent: reframing, unconditional positive regard, and solution-focused. These three attributes are derived from the representative characteristics of cognitive therapy \citep{beck-ct}, person-centered therapy \citep{rogers-pct}, and solution-focused therapy \citep{sfbt}, respectively.

In the context of cognitive therapy, cognitive reframing is referred to as cognitive restructuring. cognitive restructuring involves challenging and changing negative thoughts. It helps individuals recognize and avoid cognitive distortions—such as overgeneralization or magnifying negative behaviors—and replace them with more constructive thinking. Through this process, depressed clients can access rational thinking that is typically blocked by their cognitive distortions. In our experiment, reframing is set as an attribute to turn negative thoughts positive.

Person-centered therapy, also known as client-centered or Rogerian therapy, is based on the humanistic perspective, emphasizing the inherent worth of individuals and their potential for personal growth. 
Unconditional positive regard is a core principle and one of the three foundational conditions for therapeutic change, along with empathy and congruence in person-centered therapy. Unconditional positive regard refers to the therapist's acceptance and support of a client regardless of what the client says or does. This acceptance is not contingent on the client's behavior or characteristics, meaning it is "unconditional." In our experiment, unconditional positive regard attribute functions to provide respect, recognition, and empathy towards users.

Solution-focused brief therapy is a short-term, goal-oriented therapeutic approach that focuses on helping clients find solutions to their problems rather than dwelling on the problems themselves. In our experiment, the solution-focused attribute is used to provide the best method for users to quickly solve problems.

\section{TherapyTalk Dataset Creation}
\label{91-Dataset-Creation}
We design a new support response dataset TherapyTalk annotated by human mental health experts. We initially utilized Reddit Mental Health Dataset \citep{reddit-mental-health-dataset} comprised of user posts from subreddits related to psychological concerns. This dataset was then processed using a similarity-based methodology to structure multiple posts into cohesive units. Subsequently, mental health professionals, drawing upon their extensive background and domain knowledge, manually crafted motivational responses. This intricate process was applied to a total of 97 user posts.

\begin{enumerate}
    \item \textbf{Post Collecting} Reddit Mental Health Dataset \citep{reddit-mental-health-dataset} contains a wealth of posts related to various concerns. We sample some of these posts to use as the basis for our dataset.
 
    \item \textbf{Post Filtering} For our experiment, posts should contain sufficient lengths of content related to mental health problems, including personal anecdotes or reflections on oneself. Thus, we filtered posts to be within a range of 500 to 2,000 characters each.

    \item \textbf{Post Clustering} To create a dataset with a wider and more diverse range of information, we group multiple posts into a single example. To ensure that unrelated posts are not combined, we utilize clustering techniques. We utilize MentalRoberta \citep{mental-bert} to map the content of each post into the embeddings, and subsequently cluster them according to the K-nearest neighbor technique.

    \item \textbf{Example Composition} We randomly select one post and then identify the ten closest candidate posts. From these candidate posts, we randomly choose two, treating a total of three posts as a single example.

    \item \textbf{Example Validation} Despite grouping posts based on similarity, conflicts in the details of each post's content may still arise. To prevent this, we employ five human annotators who are experts in psychology and distribute the dataset among them. Four are working as professional counseling therapists, and one is a graduate student in psychology. They label any inconsistencies between posts. We remove all invalid examples.

    \item \textbf{Response Tagging} We instruct the experts to write responses to valid examples. We guide them to compose empathetic and helpful answers, tailored to reflect the unique characteristics of the virtual user.

    \item \textbf{Attribute Estimation} We employed other annotators to label how much the responses utilize the attributes of reframing, regard, and solution. These labels inform which method the experts select and how much the counseling strategy should be applied for each post. The attribute scores are used to develop the scoring model. To capture a broader relationship between responses and scores, attribute scores are also labeled for artificially generated responses to each post.

\end{enumerate}

Finally, each example consists of three posts containing similar concerns, and tailored and empathetic responses from experts that can provide assistance. Through this process, we construct a dataset comprising a total of 97 examples. We utilize this dataset as our test set for evaluation.

\begin{figure*}[hp]
    \centering
    \begin{tabular}{lp{0.85\linewidth}}
    \toprule 
        \textbf{Post 1} & 
        I often feel inferior to those around me. When I'm with my friends, I don't feel cool enough. This feeling is particularly strong when I talk to girls; I always feel inadequate, which causes me to pull away and stop interacting. I've never been good at talking or making conversation, leading to awkward situations, which I believe is the root of the problem. I'm not sure what I hope to gain from sharing this, but I thought some of you might relate.
        \\
        \midrule 
        \textbf{Post 2} &
        Is it common to feel like you have different personalities with different people? I’m a 19-year-old male, and lately, I've noticed that I unknowingly project different aspects of myself to different people. It's making me feel like I have split personalities. For example, some friends believe I spend all day playing video games and partying, while others think I'm deeply into art. Some see me as very introverted, while others view me as outgoing. I feel like no one truly understands who I am except for myself, and each friend group only sees certain sides of me. This is really starting to affect me, as I'm unsure of who I really am or what I want when I'm alone. Is this normal?
        \\
        \midrule
        \textbf{Post 3} &
        Socializing and interacting with others feels overwhelming and drains my energy. Do others find it easier than I do, like they're playing a complex game while I struggle with something much simpler? No matter how I approach it, I often feel like people see me as strange or unconventional. I find it confusing, whether it's initiating conversations or navigating the complexities of social norms like when it's appropriate to invite someone based on various factors. It's frustrating and it's starting to really get to me. All I want is to feel comfortable talking to people.
        \\
        \midrule
        \textbf{Response} &
        I understand how "being social" can be a lot of stress as we often have to switch on and off our personal aspects regardless of how we feel at that moment. There are also many things we have to consider before, while, and sometimes even after a conversation with others. I feel this thought overflows in our mind not just when a person is more likely to be an inferior type, but when a person loses himself or herself in a conversation. We are often way too sensitive about how other people portray us - we want to be "good". We are then slowly encroached by the thought, 'Who am I?' I know it is not easy to refrain from this chain of thought, but remember that it is actually your brain that judges you first, not the others!
        \\
        \bottomrule
    \end{tabular}
    \caption{An example of our dataset TherapyTalk. The contents of the user post are not original and are rephrased.}
    \label{fig:therapy-talk-example}
\end{figure*}

\cref{fig:therapy-talk-example} is an example from TherapyTalk.
The response follows our guidelines by incorporating information specific to each post. It includes emotional empathy towards the author and provides simple tips to help improve the author's issues.

\section{Human Evaluation Guideline and Criteria}
\label{92-Human-Evaluation}
We employ three human annotators for the evaluation. These annotators  possess knowledge of psychology. We ask them to rate the responses on a scale of 1 to 5 based on the following criteria:
\begin{itemize}
    \item \textbf{Understanding} measuring how well the response interprets and reflects the user's emotional state and experiences.
    \item \textbf{Relevance} assessing the pertinence of the response to the user's specific issues.
    \item \textbf{Professionalism} examining the integration of three strategies (Reframing, Solution-focused approaches, and Positive regard). 
    \item \textbf{Customization} determining the personalization of the response to the user's unique situation.
    \item \textbf{Satisfaction} rating how well the response aligns with the user’s preferences and expectations. 
\end{itemize}

We instruct the annotators about the goals of our study and evaluation, including explanations of the confidentiality measures in place. Additionally, we explain the content warnings associated with the sensitive material they will encounter, ensuring they are prepared to handle potentially distressing content responsibly. We also compensate the annotators at a rate of approximately \$12 per hour.
The explanation and analysis of the results are provided in \cref{5-Experiments-1}.

\section{User Study Guideline}
\label{99-Appendix}
We conduct a user study to evaluate our response generation framework, recruiting 18 participants from a diverse demographic including university students and working professionals to cover a broad range of personal concerns (\cref{sec:7-user-study}). Participants provide their concerns through a Google form and receive approximately four tailored responses in English within a week. They are then asked to select the response that best aligns with their preferences and expectations based on their personal criteria. Participants are compensated with approximately \$8 for their participation. We ensure that all participants are thoroughly informed about the study's objectives and the confidentiality measures in place. Content warnings are provided to prepare them for potentially sensitive material they might encounter during the evaluation. Additionally, participants agree to terms of use, which include detailed information on how their data will be utilized in the study.

\section{Additional Evaluation Results}
\label{93-Additional-Evaluation-Results}
\cref{tab:gpteval} presents the results of the LLM Evaluation. The analysis of these results and the details of the related experiments are discussed in \cref{5-Experiments-1}.
\begin{table*}[t!]
\resizebox{\linewidth}{!}{
\begin{tabular}{ll||ccccc||ccccc}
\toprule
\multicolumn{2}{l||}{ } & \multicolumn{5}{c||}{\textbf{Tailored Response Dataset}} & \multicolumn{5}{c}{\textbf{Counsel Chat Dataset}} \\
\midrule
Model & Method & Cus. & Sat. & Pro. & Rel. & Und. & Cus. & Sat. & Pro. & Rel. & Und. \\
\midrule
GPT-4-0125 & SA & 3.88 & 4.00 & 4.13 & 4.19 & 4.25 & 4.36 & 4.27 & 4.64 & 4.27 & 4.00   \\
& SAA & 3.63 & 4.00 & 4.00 & 3.88 & 4.16 & 4.09 & 4.09 & 4.36 & 4.27 & 4.00  \\
& MAA & 4.06 & 4.19 & 3.69 & 4.25 & 4.35 & 4.10 & 4.10 & 4.27 & 4.27 & 4.33  \\
& \mcraft{} & \textbf{4.69} & \textbf{4.75} & \textbf{4.25} & \textbf{4.94} & \textbf{5.00} & \textbf{4.36} & \textbf{4.55} & \textbf{4.82} & \textbf{4.73} & \textbf{4.33} \\
\midrule
GPT-3.5-turbo & SA & 3.56 & 3.78 & 4.11 & 3.89 & 3.78 & 4.36 & 4.27 & 4.64 & 4.27 & 4.00 \\
& SAA & 3.33 & 3.11 & 3.56 & 3.56 & 3.33 & 4.09 & 4.09 & 4.36 & 4.27 & 4.00 \\
& MAA & 3.78 & 3.78 & 4.33 & 4.00 & 4.00 & 4.09 & 4.09 & 4.27 & 4.27 & 4.33 \\
& \mcraft{} & \textbf{4.33} & \textbf{4.22} & \textbf{4.89} & \textbf{4.67} & \textbf{4.67} & \textbf{4.36} & \textbf{4.55} & \textbf{4.82} & \textbf{4.73} & \textbf{4.33} \\
\midrule
LLAMA-2-13b & SA & 1.70 & 1.80 & 1.90 & 1.80 & 1.90 & 3.25 & 3.50 & 3.88 & 3.50 & 3.50 \\
& SAA & 1.90 & 2.00 & 2.00 & 1.90 & 2.20 & 2.62 & 2.62 & 2.75 & 2.62 & 2.75 \\
& MAA & 1.80 & 1.80 & 1.90 & 1.80 & 1.80 & 4.00 & 4.00 & 4.00 & 4.25 & 4.25 \\
& \mcraft{} & \textbf{2.30} & \textbf{2.20} & \textbf{2.20} & \textbf{2.40} & \textbf{2.30} & \textbf{4.62} & \textbf{4.88} & \textbf{4.75} & \textbf{4.62} & \textbf{4.88} \\

\midrule

MentalAlpaca & SA & 3.92 & 4.00 & 4.08 & 4.08 & 3.92 & 3.25 & 3.50 & 3.88 & 3.50 & 3.50 \\
& SAA & 2.15 & 2.15 & 2.38 & 2.08 & 2.23 & 2.62 & 2.62 & 2.75 & 2.62 & 2.75 \\
& MAA & 3.92 & 3.77 & 4.23 & 3.92 & 4.08 & \textbf{4.62} & \textbf{4.88} & \textbf{4.75} & \textbf{4.62} & \textbf{4.88}\\
& \mcraft{} & \textbf{4.31} & \textbf{4.15 } & \textbf{4.31} & \textbf{4.38} & \textbf{4.23} & 4.00 & 4.00 & 4.00 & 4.25 & 4.25 \\
\bottomrule
\end{tabular}
}
\caption{This table showcases the scores assigned by GPT4-turbo, serving as a judge using a 5-point Likert scale across two evaluation datasets. The evaluation criteria include Customization (Cus.), Satisfaction (Sat.), Professionalism (Pro.), Relevance (Rel.), and Understanding (Und.). Scores range from 1 (Poor) to 5 (Excellent).}
\label{tab:gpteval}
\end{table*}

\section{Prompts}
\label{94-Prompts}
\begin{figure*}[t!]
    \centering
    \includegraphics[width=1\linewidth]{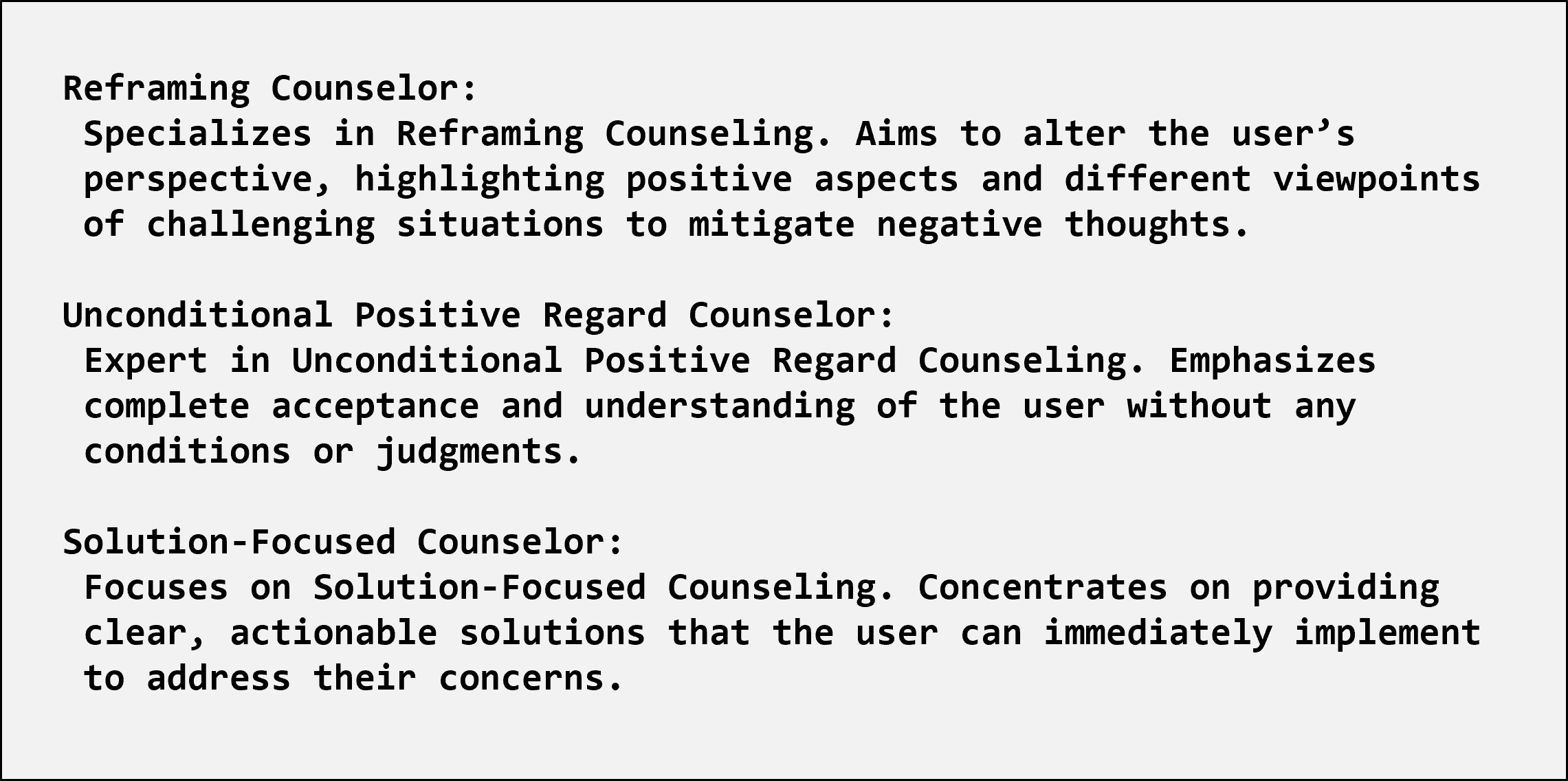}
    \caption{Prompt examples of each agent role setup using in Strategic Agents Setup of \cref{3-Method}.}
    \label{fig:role-prompt}
\end{figure*}

\begin{figure*}[t!]
    \centering
    \includegraphics[width=1\linewidth]{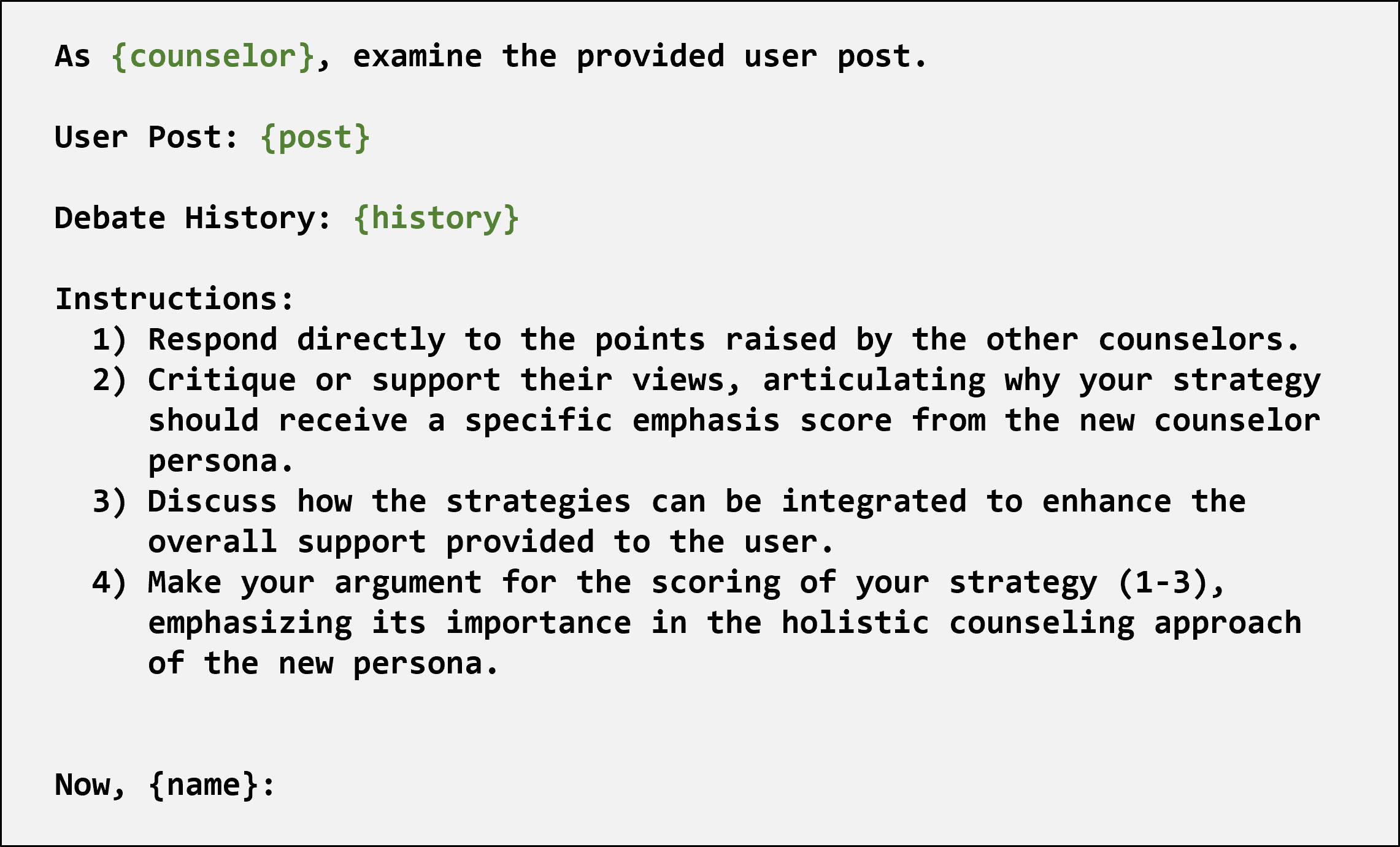}
    \caption{Prompt examples of simulating debating session between agents using in Strategic Debating of \cref{3-Method}. The green slots are replaced with actual data before generating.}
    \label{fig:debate-prompt}
\end{figure*}

 \cref{fig:role-prompt} and \cref{fig:debate-prompt} provide examples of the role-setting prompts for each agent and the prompts used for debating in our framework.

\section{Examples}
\label{96-Examples}
\cref{fig:case_study} presents a case study comparing MentalAgora with other configurations, focusing on the inclusion of specific counseling attributes. The responses are color-coded to highlight these attributes: red indicates Reframing, orange indicates Unconditional Positive Regard, and blue indicates Solution-Focused. By systematically removing each attribute from the responses, we can analyze the impact of their absence or presence. The figure is an example of the experimental results discussed in \cref{5-Experiments}. The post input is not the actual user post and is paraphrased.

\begin{table*}[t!]
\centering
{
\begin{tabular}{p{0.12\textwidth}p{0.82\textwidth}}
\toprule
\textbf{Post} & Whenever I go on a date with someone new, I get so anxious beforehand that I feel sick, and it's becoming increasingly frustrating. This started in the past year, ever since I began exploring the dating scene. I keep telling myself that it's no big deal, it's just one date, I don't have to see the person again if I don't want to, and I won't be pressured into anything. But despite these reassurances, on the day of the date, I can't control the anxiety. I get a stomach ache that feels like intense butterflies, which makes me nauseous. All I can think about is how sick I feel and how I don't want to show up to the date like this. I want to enjoy it, but I can't make the feeling stop. Normally, I manage my anxiety well in social situations, like meeting groups of people or friends of friends, but my brain seems to interpret dates differently. It's really frustrating. \\
\midrule
\midrule
\textbf{Method} & \textbf{Response} \\
\midrule
\mcraft & \textcolor{orange}{It sounds like you're facing significant anxiety with dating, a common challenge yet uniquely intense for each person.} Your capacity to handle social situations well typically, yet feeling overwhelmed by dating, highlights how personal and intricate anxiety triggers can be. Seeing these feelings as your body's way of signaling a deep desire for connection and personal growth can be empowering. Anxiety, in its essence, isn't your enemy but a part of you that's asking for attention and care.
\textcolor{red}{Let's try reframing this experience: What if these intense feelings are not barriers but signposts, guiding you towards understanding what you truly seek in relationships and within yourself? This perspective doesn't downplay your discomfort but acknowledges your strength and resilience in facing these emotions.}
\textcolor{blue}{Practically speaking, before your next date, try this strategy: Allocate a specific time to focus on your concerns, say 10 minutes, to acknowledge and address them. Then, shift your focus to positive affirmations and visualize a successful and enjoyable date experience.} \\
\midrule
Removing Reframing & It sounds like you've found yourself in a really tough spot, where your excitement for new connections is overshadowed by substantial anxiety. \textcolor{orange}{It's perfectly okay to feel this way, and your experiences are valid. Dating can indeed trigger deeper anxieties, and it's commendable that you're seeking ways to navigate this.} Together, let's explore strategies to help manage these feelings. \textcolor{blue}{First, consider grounding techniques before your dates, such as deep breathing or visualizing a positive outcome to help ease your nervous system. It might also be beneficial before the date to engage in an activity you find relaxing, whether it's listening to music, a short walk, or reading something enjoyable. Additionally, setting smaller, manageable goals for each date can gradually build your confidence.} \\
\midrule
Removing Positive-regard & It sounds like you're experiencing intense anxiety specific to dating, which is both distressing and frustrating, especially when you feel confident in other social situations. \textcolor{blue}{Let's explore two paths to manage this: first, practical steps to reduce anxiety in the moment, and second, shifting your perspective on dating to reduce long-term stress. For immediate relief, consider establishing a pre-date routine that calms you. This could be deep breathing exercises, a short walk, listening to your favorite music, or anything else that grounds you. Additionally, setting up a safe call or text with a friend during or after the date can provide a comforting safety net.} \textcolor{red}{In terms of reframing your perspective, it might help to see each date not as an evaluation of your worth.}\\
\bottomrule
\end{tabular}
}
\caption{Case study of \textit{MentalAgora} and other configurations. The responses highlight the inclusion of specific counseling attributes: \textcolor{red}{red} indicates \textit{Reframing}, \textcolor{orange}{orange} indicates \textit{Unconditional Positive Regard}, and \textcolor{blue}{blue} indicates \textit{Solution-Focused}. By removing each attribute from the responses, we can observe how the absence or presence of each attribute impacts the overall response. For instance, in the absence of \textit{Reframing}, the response may lack a shift in perspective, while the absence of \textit{Unconditional Positive Regard} might result in less empathy and acceptance. Conversely, some responses might inadvertently emphasize other attributes more strongly when one is removed.}
\label{fig:case_study}
\end{table*}

\begin{table*}[t!]
\centering
{
\begin{tabular}{p{0.12\textwidth}p{0.82\textwidth}}
\toprule
Removing Solution-focused & \textcolor{orange}{It sounds like you've encountered quite a challenging hurdle in your dating experiences, and it's understandable how this level of anxiety can feel overwhelming.} What stands out is your awareness and introspection about your feelings. in your dating experiences, and it's understandable how this level of anxiety can feel overwhelming. What stands out is your awareness and introspection about your feelings. That's a significant first step. With the nervousness manifesting physically, it's clear this response isn't something you can simply talk yourself out of, and that's okay. Your feelings are valid, and it's important to acknowledge them without judgment. \textcolor{red}{Incorporating a blend of acceptance and cognitive reframing might offer some relief. Begin by affirming that it's perfectly normal to feel anxious about dating—it's a sign of your desire for meaningful connections. Then, let's gently explore these feelings. What specific thoughts contribute to your anxiety? Can we challenge them or look at them differently?  Maybe we can gradually shift the perspective from fearing.} \\
\midrule
Reframing only & It sounds like you're experiencing intense pre-date anxiety, and I hear how frustrating and challenging this is for you. Your feelings are valid, and it's common to feel anxious about dating. \textcolor{red}{Consider trying to reframe your thoughts about dating. Instead of viewing each date as a high-stakes event, think of it as an opportunity to meet someone new and learn about them, with no pressure on the outcome.} To manage your physiological symptoms, deep breathing exercises can be helpful. Try taking deep, slow breaths before your date to calm your nervous system. Additionally, \textcolor{red}{grounding techniques, like noticing five things you can see, four things you can touch, three things you can hear, two things you can smell, and one thing you can taste, can also bring you back to the present moment and help reduce anxiety.} \\
\midrule
Positive-regard only & It sounds like you're experiencing a lot of anxiety around dating, which is understandable given it's a relatively new experience for you. \textcolor{orange}{Your feelings are completely valid, and it's important to remember you're not alone in feeling this way.} Dating can indeed be nerve-wracking, but there are ways to manage the anxiety you're feeling. First, applying Cognitive-Behavioral Techniques (CBT) might help you challenge and reframe any negative thoughts you have about dating. For instance, instead of thinking, "This date will go poorly because I'm anxious," try to think, \textcolor{orange}{"I'm anxious because I'm trying something new, and that's okay. I can handle it. Incorporating Mindfulness-Based Stress Reduction (MBSR) techniques can also be beneficial.}\\
\midrule
Solution-focused only  & I understand how frustrating and challenging it can feel to experience such intense anxiety before a date, especially when you're eager to enjoy these experiences. The physical symptoms you're describing are common manifestations of anxiety, and it's important to acknowledge them while also working towards managing their impact. \textcolor{blue}{Let's start by addressing the immediate symptoms with some practical strategies. Deep breathing exercises, mindfulness, or even a brief walk can be helpful to soothe your nervous system. Also, preparing a small `ritual' before your dates—like listening to a favorite song or having a reassuring talk with a friend—can serve as a grounding mechanism. Now, for a more lasting strategy, consider reflecting on past situations where you've successfully managed your anxiety.} What skills did you utilize then that could be applicable now? Incrementally setting goals for each date can also build your confidence.\\
\midrule
\bottomrule
\end{tabular}
}
\caption{Case study of \textit{MentalAgora} and other configurations. The responses highlight the inclusion of specific counseling attributes: \textcolor{red}{red} indicates \textit{Reframing}, \textcolor{orange}{orange} indicates \textit{Unconditional Positive Regard}, and \textcolor{blue}{blue} indicates \textit{Solution-Focused}. By removing each attribute from the responses, we can observe how the absence or presence of each attribute impacts the overall response. For instance, in the absence of \textit{Reframing}, the response may lack a shift in perspective, while the absence of \textit{Unconditional Positive Regard} might result in less empathy and acceptance. Conversely, some responses might inadvertently emphasize other attributes more strongly when one is removed.}
\label{tab:case_study}
\end{table*}

\end{document}